\documentclass{article}
\usepackage[preprint]{neurips_2024}
\usepackage[utf8]{inputenc}
\usepackage[T1]{fontenc}
\usepackage{hyperref}
\usepackage{url}
\usepackage{booktabs}
\usepackage{amsfonts}
\usepackage{nicefrac}
\usepackage{microtype}
\usepackage{xcolor}
\usepackage{tabularx}
\usepackage{amsmath, amssymb, amsthm}
\usepackage{mathtools}
\usepackage{graphicx}
\usepackage{multirow}
\usepackage{tikz}
\usetikzlibrary{arrows.meta,positioning,fit,backgrounds,calc,shapes.geometric}

\graphicspath{{../../results/}{figures/}{../../assets/}}

\definecolor{figblue}{HTML}{2E5EAA}
\definecolor{figbluefill}{HTML}{EAF1FB}
\definecolor{figorange}{HTML}{C56A12}
\definecolor{figorangefill}{HTML}{FDF0E3}
\definecolor{figpurple}{HTML}{6D4AA5}
\definecolor{figpurplefill}{HTML}{F2ECFA}
\definecolor{figgray}{HTML}{5D6673}
\definecolor{figgrayfill}{HTML}{F4F6F8}
\definecolor{figred}{HTML}{B03A2E}
\definecolor{figgreen}{HTML}{2E7D32}

\newtheorem{theorem}{Theorem}[section]

\newtheorem{proposition}[theorem]{Proposition}
\newtheorem{corollary}[theorem]{Corollary}
\newtheorem{definition}[theorem]{Definition}
\newtheorem{assumption}[theorem]{Assumption}

\DeclareMathOperator{\Red}{Red}
\DeclareMathOperator{\Unq}{Unq}
\DeclareMathOperator{\Syn}{Syn}
\DeclareMathOperator{\Cov}{Cov}

\title{Asymmetric Multimodal Learning: The Alignment Redundancy Bound and Decomposed JEPAs}

\author{%
  Yusuf Shihata \\
  \texttt{yusufshihata2006@gmail.com}
}

\begin{document}

\maketitle

\begin{abstract}
Cross-modal alignment is the default objective for multimodal representation
learning, but it rests on an implicit assumption: that the two modalities are
roughly symmetric views of the same underlying state. We show that this
assumption fails on real multimodal data, and that alignment as a sole
objective is provably lossy under informational asymmetry. We formalize the
failure through Partial Information Decomposition, proving that cross-modal
alignment retains at most the redundant information atom and annihilates the
modality-unique and synergistic structure on which asymmetric tasks depend. We
give the exact PID decomposition of the alignment information bottleneck,
characterize its failure regime, and derive a sufficiency condition that any
encoder objective must satisfy to preserve all four informational atoms. We
then introduce Decomposed JEPA (D-JEPA), which realizes this condition by
factorizing each modality's latent space into an aligned shared core and a
private fiber anchored by contextual masked prediction. On compositional
reasoning benchmarks, alignment-only training undergoes catastrophic semantic
collapse on relational queries; D-JEPA prevents this collapse, recovering the
referent structure alignment discards, restoring most of the mutual information
available from a frozen backbone, and producing positive multimodal synergy
where alignment inverts it. Information-theoretic and causal audits confirm
that the recovered information is causally localized in the private fibers.

\end{abstract}

\section{Introduction}
\label{sec:intro}

Cross-modal alignment is the default objective for building multimodal
representations. Its appeal is not merely practical: it rests on a specific and
usually implicit claim about the data. When two observations $X_1$ and $X_2$ are
equally informative views of the same underlying state $Z$---two simultaneous
camera angles of one scene, or a dense image paired with an exhaustive
caption---the task-relevant content of each view is content the other also
carries, and forcing their latents to agree is the right thing to do. We call
this the \textbf{redundancy premise}: that what matters about $Z$ is shared
across the modalities that observe it. Under the redundancy premise, alignment
is not merely a convenient proxy for shared structure; it is the objective that
recovers it, and anything alignment discards is, by construction, not worth
keeping.

The premise is rarely true, and it is rarely checked. Real multimodal pairs
satisfy it only when the pairing is dense and descriptive. Consider visual
question answering: the image $X_1$ encodes a dense field of objects, textures,
and spatial relations, while the question $X_2$ encodes a selective relational
operator (e.g., \emph{``how many blue cubes are behind the cylinder?''}). The
question does not describe the scene; it interrogates it. Much of what the image
contains is unexpressible in the query, and much of what the query asserts is
nowhere in the image. The two modalities are \textbf{informationally asymmetric}:
neither can express what the other carries, and the information the task
requires is not the information the pair shares.

When an alignment objective is applied to such a pair, it forces the dense
observation and the selective query to predict each other in latent space.
Because Dirichlet energy minimization treats any feature absent from \emph{both}
modalities as noise, alignment discards the specific linguistic referent from
$X_2$ and the fine-grained visual structure from $X_1$ that no query names. This
is not a small correction to an otherwise sound objective. It is the objective
working exactly as specified, on data the specification was not meant for. The
redundancy premise is not approximately violated here; it is inverted, and the
loss is concentrated precisely in the structure the task depends on.

We therefore ask: \textbf{what information does alignment discard when modalities
are informationally asymmetric, and what architecture retains it?}

\begin{figure}[t]
\centering
\resizebox{\textwidth}{!}{%
\begin{tikzpicture}[
  font=\sffamily\small,
  inp/.style={draw=figgray!55, fill=white, rounded corners=3pt, font=\scriptsize, inner sep=4pt},
  opb/.style={draw=figblue!80!black, fill=figbluefill, rounded corners=4pt, font=\scriptsize, align=center, inner sep=5pt},
  opp/.style={draw=figpurple!80!black, fill=figpurplefill, rounded corners=4pt, font=\scriptsize, align=center, inner sep=5pt},
  nb/.style={draw=figblue!80!black, fill=figbluefill, rounded corners=3pt, font=\scriptsize, align=center, inner sep=4pt},
  no/.style={draw=figorange!85!black, fill=figorangefill, rounded corners=3pt, font=\scriptsize, align=center, inner sep=4pt},
  np/.style={draw=figpurple!80!black, fill=figpurplefill, rounded corners=3pt, font=\scriptsize, align=center, inner sep=4pt},
  nd/.style={draw=figred!70, dashed, fill=figgrayfill, rounded corners=3pt, font=\scriptsize, align=center, inner sep=4pt},
  arr/.style={-{Latex[length=1.6mm]}, draw=figgray!70, line width=0.7pt},
  arrb/.style={arr, draw=figblue!80!black},
  arro/.style={arr, draw=figorange!85!black},
  arrp/.style={arr, draw=figpurple!80!black},
  arrr/.style={-{Latex[length=1.6mm]}, draw=figred, dashed, line width=0.7pt},
  bus/.style={draw=figgray!70, line width=0.7pt},
]
  \begin{scope}[local bounding box=panelA]
    \fill[figbluefill!25, rounded corners=8pt] (0,0) rectangle (7.0,5.0);
    \draw[figgray!25, line width=0.6pt, rounded corners=8pt] (0,0) rectangle (7.0,5.0);
    \node[anchor=west, font=\scriptsize\bfseries, text=figgray!90] at (0.30,4.65) {(a) Alignment-only (symmetric assumption)};

    \node[inp] (x1) at (0.95,3.55) {Vision $X_1$};
    \node[inp] (x2) at (0.95,2.05) {Text $X_2$};
    \node[opb] (opA) at (3.10,2.80) {Alignment Method};
    \node[nb] (plat) at (6.00,3.55) {Shared $z^\star$\\{\tiny ($\Red$ only)}};
    \node[nd] (discard) at (6.00,2.05) {$\Unq_1, \Unq_2$\\$\Syn$\\{\tiny\color{figred}\bfseries Destroyed}};

    \draw[arr] (x1) -- (opA);
    \draw[arr] (x2) -- (opA);
    \draw[arr] (opA.east) -- (4.63,2.80);
    \draw[bus] (4.63,2.05) -- (4.63,3.55);
    \draw[arrb] (4.63,3.55) -- (plat.west);
    \draw[arrr] (4.63,2.05) -- (discard.west);

    \node[anchor=west, font=\tiny\bfseries, text=figgray] at (0.35,1.30) {Information Retained:};
    \fill[figblue!80!black, rounded corners=2pt] (0.35,0.50) rectangle (2.15,0.92);
    \node[font=\scriptsize\bfseries, text=white] at (1.25,0.71) {Redundancy};
    \fill[figgrayfill!70, rounded corners=2pt] (2.35,0.50) rectangle (6.65,0.92);
    \draw[figgray!45, dashed, rounded corners=2pt] (2.35,0.50) rectangle (6.65,0.92);
    \node[font=\scriptsize, text=figgray!85] at (4.50,0.71) {Discarded: $\Unq_1, \Unq_2, \Syn$};
  \end{scope}

  \begin{scope}[local bounding box=panelB, xshift=7.5cm]
    \fill[figpurplefill!25, rounded corners=8pt] (0,0) rectangle (7.0,5.0);
    \draw[figgray!25, line width=0.6pt, rounded corners=8pt] (0,0) rectangle (7.0,5.0);
    \node[anchor=west, font=\scriptsize\bfseries, text=figgray!90] at (0.30,4.65) {(b) Decomposed JEPA (D-JEPA)};

    \node[inp] (bx1) at (0.95,3.55) {Vision $X_1$};
    \node[inp] (bx2) at (0.95,2.05) {Text $X_2$};
    \node[opp] (decomp) at (3.10,2.80) {D-JEPA};

    \node[nb] (c)   at (6.00,3.95) {$c$ ($\Red$)};
    \node[no] (u1)  at (6.00,3.15) {$u_1$ ($\Unq_1$)};
    \node[no] (u2)  at (6.00,2.35) {$u_2$ ($\Unq_2$)};
    \node[np] (syn) at (6.00,1.55) {Synergy\\{\tiny ($\Syn$)}};

    \draw[arr] (bx1) -- (decomp);
    \draw[arr] (bx2) -- (decomp);
    \draw[arr] (decomp.east) -- (4.63,2.80);
    \draw[bus] (4.63,1.55) -- (4.63,3.95);
    \draw[arrb] (4.63,3.95) -- (c.west);
    \draw[arro] (4.63,3.15) -- (u1.west);
    \draw[arro] (4.63,2.35) -- (u2.west);
    \draw[arrp] (4.63,1.55) -- (syn.west);

    \node[anchor=west, font=\tiny\bfseries, text=figgray] at (0.35,1.30) {Information Retained:};
    \fill[figblue!80!black, rounded corners=2pt] (0.35,0.50) rectangle (1.775,0.92);
    \node[font=\scriptsize\bfseries, text=white] at (1.0625,0.71) {$\Red$};
    \fill[figorange!85!black, rounded corners=2pt] (1.975,0.50) rectangle (3.40,0.92);
    \node[font=\scriptsize\bfseries, text=white] at (2.6875,0.71) {$\Unq_1$};
    \fill[figorange!70!black, rounded corners=2pt] (3.60,0.50) rectangle (5.025,0.92);
    \node[font=\scriptsize\bfseries, text=white] at (4.3125,0.71) {$\Unq_2$};
    \fill[figpurple!80!black, rounded corners=2pt] (5.225,0.50) rectangle (6.65,0.92);
    \node[font=\scriptsize\bfseries, text=white] at (5.9375,0.71) {$\Syn$};
  \end{scope}
\end{tikzpicture}%
}
\caption{\textbf{Alignment under informational asymmetry.} (a) When two
modalities are not redundant views of one state (e.g., scene context vs.\
relational query), cross-modal invariance projects both onto their redundant
intersection $\Red$, destroying unique referents $\Unq_m$ and synergistic
bindings $\Syn$. (b) D-JEPA factorizes the latent space into an aligned shared
core $c$ and modality-specific private fibers $u_m$, anchored by contextual
masked prediction.}
\label{fig:teaser}
\end{figure}
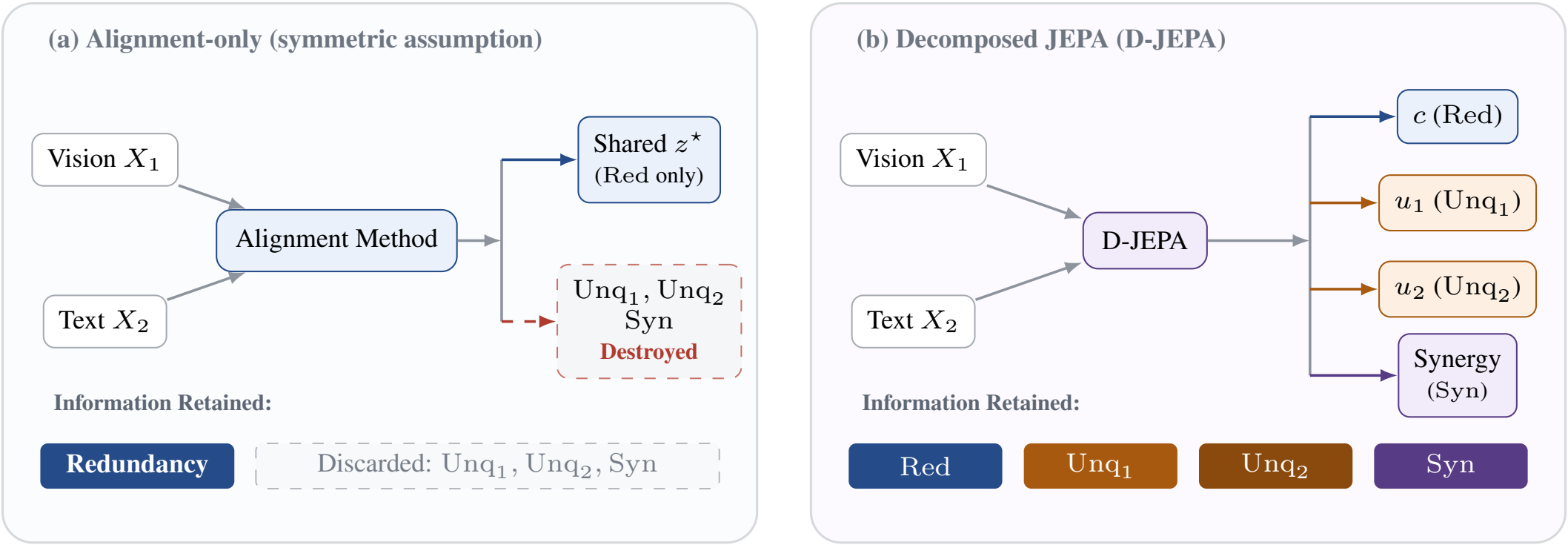

Through the lens of Partial Information Decomposition
(PID)~\citep{williams2010nonnegative,bertschinger2014quantifying}, we make four
contributions:

\begin{enumerate}
    \item \textbf{The Alignment Redundancy Bound.} We prove that any optimal
    alignment-only method captures at most the \emph{redundant} information atom
    $\Red(Z; X_1, X_2)$. The unique components $\Unq(Z; X_1), \Unq(Z; X_2)$ and
    the synergistic component $\Syn(Z; X_1, X_2)$ are annihilated by
    cross-modal invariance (Theorem~\ref{thm:alignment_redundancy}). The bound
    is a statement about the alignment objective alone: it follows from
    Dirichlet energy minimization over conditionally dependent, asymmetric
    views, and assumes nothing about how representations might converge absent
    that objective.

    \item \textbf{Theoretical bound and empirical collapse.} We prove that
    complete cross-modal invariance bounds retained information by the
    redundant atom, and demonstrate empirically that under asymmetric
    task-relevant structure this manifests as dimensional collapse: the model
    suppresses fine-grained linguistic and visual referents to predict generic
    centroids. On CLEVR, an alignment-only model falls to $0.43\%$ on
    attributes and $4.88\%$ on counting, inverting the multimodal synergy gap
    to $-1.56\,\text{pp}$.

    \item \textbf{Decomposed JEPA (D-JEPA).} We introduce an architecture that
    factorizes representations into an aligned shared core ($\mathcal{C}$) and
    private modality-specific fibers ($\mathcal{U}_m$). Factorization escapes the
    redundancy bound by construction, combining scale-invariant cosine Dirichlet
    alignment, contextual target-backbone prediction, and an asymmetric covariance
    penalty ($\operatorname{Cov}(u_m, c_m) = 0$) that prevents leakage without
    suppressing cross-modal coordination.

    \item \textbf{Empirical auditing on diagnostic benchmarks.} On CLEVR, D-JEPA reaches $\mathbf{39.61 \pm 3.07\%}$
    downstream accuracy against $21.25 \pm 0.69\%$ for alignment-only ($\mathbf{+18.36\,\text{pp}}$,
    $95\%$ paired bootstrap CI $[17.21, 19.44]\,\text{pp}$) and $27.96\%$ for the same baseline
    augmented with reconstruction ($\mathbf{+11.65\,\text{pp}}$). Probe-PID and Flow-PID audits
    show D-JEPA restores mutual information from $0.24$ to $1.94$ bits ($84.5\%$ of the
    frozen-backbone ceiling), and causal fiber-swap interventions demonstrate that the
    recovered information is causally localized in the private fibers.
\end{enumerate}

\paragraph{Code and reproducibility.}
The reference implementation, the synthetic-data generator, the CLEVR
preprocessing pipeline, the probe and PID audit scripts, and the
figure-generation code for every table and figure in this paper are
available at \url{https://github.com/Vixel2006/D-JEPA}.

\section{Theory}
\label{sec:theory}

We formalize multimodal representation learning from a generative world-state perspective, prove that alignment-only bottlenecks retain only redundant information, establish sufficiency conditions for recovering the full world state, and characterize what architectural fusion can extract from general encoder representations.

\subsection{Setup and Notation}

Let $S \sim P(S)$ denote the underlying world state, and let $\mathcal{M}_1, \mathcal{M}_2$ be two modality spaces with observations $X_i \in \mathcal{M}_i$ ($i \in \{1, 2\}$) drawn from $P(X_1, X_2 \mid S)$. We consider modality encoders $q_i(z_i \mid x_i)$ mapping each observation $X_i$ to a representation $Z_i \in \mathbb{R}^k$ ($k > 0$)---for instance, $Z_i = f_i(X_i) + \varepsilon_i$ for a smooth map $f_i: \mathcal{M}_i \to \mathbb{R}^k$ and small noise $\varepsilon_i$, ensuring finite mutual information while satisfying the local Markov condition $Z_i \perp (S, X_{\neq i}) \mid X_i$.

\begin{definition}[Encoder Information Decomposition]
\label{def:causal_decomp}
For an encoder $Z_i$ satisfying $Z_i \perp X_{\neq i} \mid X_i$, the mutual information between the representation $Z_i$ and its input $X_i$ decomposes into
\begin{equation}
\label{eq:causal_decomp}
    I(Z_i; X_i) = I(Z_i; X_i \mid X_{\neq i}) + I(Z_i; X_i; X_{\neq i}).
\end{equation}
\end{definition}

By symmetry of interaction information and the Markov condition $I(Z_i; X_{\neq i} \mid X_i) = 0$, the second term simplifies to
\begin{equation}
    I(Z_i; X_i; X_{\neq i}) = I(Z_i; X_{\neq i}) - \underbrace{I(Z_i; X_{\neq i} \mid X_i)}_{=\,0} = I(Z_i; X_{\neq i}).
\end{equation}
Equivalently, applying Partial Information Decomposition (PID)~\citep{williams2010nonnegative,bertschinger2014quantifying} with target $Z_i$ and sources $(X_i, X_{\neq i})$---where $I(Z_i; X_i, X_{\neq i}) = I(Z_i; X_i)$ implies $\Unq(Z_i; X_{\neq i}) = \Syn(Z_i; X_i, X_{\neq i}) = 0$---yields the final form of Definition~\ref{def:causal_decomp}:
\begin{equation}
\label{eq:encoder_pid}
    I(Z_i; X_i) = \underbrace{I(Z_i; X_i \mid X_{\neq i})}_{\Unq(Z_i;\, X_i)} + \underbrace{I(Z_i; X_{\neq i})}_{\Red(Z_i;\, X_i, X_{\neq i})}.
\end{equation}

While~\eqref{eq:encoder_pid} separates what $Z_i$ encodes exclusively from $X_i$ versus what is shared with $X_{\neq i}$, recovering the world state $S$ requires examining PID with $S$ as the target:
\begin{equation}
\label{eq:world_pid}
    I(S; X_1, X_2) = \Red(S; X_1, X_2) + \Unq(S; X_1) + \Unq(S; X_2) + \Syn(S; X_1, X_2).
\end{equation}
Informational asymmetry encompasses regimes where $\Unq(S; X_i)$ and $\Syn(S; X_1, X_2)$ are non-negligible relative to $\Red(S; X_1, X_2)$---including both capacity asymmetry ($\Unq(S; X_1) \gg \Unq(S; X_2)$) and operator--field asymmetry ($\Red \approx 0$ with $\Syn$ dominant).

\subsection{What the Alignment Information Bottleneck Optimizes}

Consider the standard Alignment Information Bottleneck (AIB) objective, which maximizes cross-modal dependency while compressing each modality's encoding:
\begin{equation}
\label{eq:aib_obj}
    \mathcal{J}_{\mathrm{AIB}}(Z_1, Z_2) \;=\; I(Z_1; Z_2) - \beta \sum_{i=1}^{2} I(Z_i; X_i), \qquad \beta > 0.
\end{equation}

\begin{assumption}[Conditional independence of views]
\label{ass:markov}
$X_1 \perp X_2 \mid S$, so local encoding yields the Markov chain $Z_1 - X_1 - S - X_2 - Z_2$.
\end{assumption}

\begin{theorem}[Alignment Redundancy Bound on the AIB Objective]
\label{thm:alignment_redundancy}
Under Assumption~\ref{ass:markov} and the Williams--Beer PID axioms applied with target $S$ and sources $(Z_1, Z_2)$, the AIB objective~\eqref{eq:aib_obj} equals
\begin{align}
\label{eq:aib_pid_exact}
    \mathcal{J}_{\mathrm{AIB}}(Z_1, Z_2)
    &= (1 - 2\beta)\,\Red(S; Z_1, Z_2) \nonumber \\
    &\quad - \Syn(S; Z_1, Z_2) \nonumber \\
    &\quad - \beta \sum_{i=1}^{2} \Unq(S; Z_i) \nonumber \\
    &\quad - \beta \sum_{i=1}^{2} I(Z_i; X_i \mid S).
\end{align}
Consequently, for any $\beta \in (0, \tfrac{1}{2})$, maximizing $\mathcal{J}_{\mathrm{AIB}}$ explicitly maximizes world-state redundancy $\Red(S; Z_1, Z_2)$ while simultaneously minimizing modality-unique information $\Unq(S; Z_i)$, cross-modal synergy $\Syn(S; Z_1, Z_2)$, and world-irrelevant noise $I(Z_i; X_i \mid S)$. For $\beta \ge \tfrac{1}{2}$, all coefficients are non-positive and the representations collapse.
\end{theorem}

\begin{corollary}[Annihilation of unique and synergistic structure]
\label{cor:annihilation}
Under Theorem~\ref{thm:alignment_redundancy}, any AIB solution constrained by $I(Z_i; X_i \mid X_{\neq i}) \to 0$ (or exact cross-modal alignment $Z_1 = Z_2$ a.s.) satisfies $I(S; Z_1, Z_2) \le \Red(S; X_1, X_2)$, discarding at least
\begin{equation}
    I(S; X_1, X_2) - I(S; Z_1, Z_2)
    \;\ge\; \Unq(S; X_1) + \Unq(S; X_2) + \Syn(S; X_1, X_2).
\end{equation}
\end{corollary}

The proofs of Theorem~\ref{thm:alignment_redundancy} and Corollary~\ref{cor:annihilation} are given in Appendix~\ref{app:proofs}.

\subsection{Joint World-State Sufficiency via $I(Z_i; X_i)$}

Theorem~\ref{thm:alignment_redundancy} shows that penalizing $I(Z_i; X_i)$ against $I(Z_1; Z_2)$ eliminates $\Unq(S; Z_i)$ and $\Syn(S; Z_1, Z_2)$. We now establish the converse: maximizing $I(Z_i; X_i)$ closes the world-state sufficiency gap and preserves all four PID atoms for any downstream task.

\begin{theorem}[Joint World-State Sufficiency]
\label{thm:sufficiency}
Let $Z_i \sim q_i(z_i \mid x_i)$ satisfy $Z_i \perp (S, X_{\neq i}, Z_{\neq i}) \mid X_i$, and let $T = h(S)$ be any downstream task determined by the world state $S$. Then:
\begin{enumerate}
    \item \textbf{Exact Sufficiency Gap:} The world-state information lost by $(Z_1, Z_2)$ equals
    \begin{align}
    \label{eq:suff_gap}
        I(S; X_1, X_2) - I(S; Z_1, Z_2) &= I(S; X_1, X_2 \mid Z_1, Z_2).
    \end{align}
    For discrete (or quantized) observations $X_i$, maximizing $I(Z_i; X_i)$ directly closes this gap:
    \begin{align}
    \label{eq:discrete_gap_bound}
        I(S; X_1, X_2) - I(S; Z_1, Z_2) &\;\le\; \sum_{i=1}^{2} \big(H(X_i) - I(Z_i; X_i)\big),
    \end{align}
    and for any task $T = h(S)$, $I(T; X_1, X_2) - I(T; Z_1, Z_2) \le I(S; X_1, X_2 \mid Z_1, Z_2)$.
    \item \textbf{Full PID Preservation:} Whenever $(Z_1, Z_2)$ closes both marginal and joint sufficiency gaps ($I(S; X_i \mid Z_i) = 0$ for $i \in \{1,2\}$ and $I(S; X_1, X_2 \mid Z_1, Z_2) = 0$), all four PID atoms of $(Z_1, Z_2)$ about $S$ equal those of $(X_1, X_2)$:
    \begin{align}
        \Red(S; Z_1, Z_2) &= \Red(S; X_1, X_2), \nonumber \\
        \Unq(S; Z_i) &= \Unq(S; X_i), \nonumber \\
        \Syn(S; Z_1, Z_2) &= \Syn(S; X_1, X_2).
    \end{align}
\end{enumerate}
\end{theorem}

The proof is given in Appendix~\ref{app:proofs}.

\subsection{Architectural Alignment via Cross-Attention Fusion}
\label{sec:fusion_theory}

Theorems~\ref{thm:alignment_redundancy} and~\ref{thm:sufficiency} characterize what encoder representations $(Z_1, Z_2)$ retain. We now analyze what a general \textbf{fusion architecture} can extract from $(Z_1, Z_2)$ without changing the encoder objective. Consider a standard cross-attention fusion block $\Phi_{\mathrm{attn}}(Z_1, Z_2)$ acting on token-level representations $Z_1 = (Z_{1,1}, \dots, Z_{1,N_1})$ and $Z_2 = (Z_{2,1}, \dots, Z_{2,N_2})$, where one modality supplies queries $Q = W_Q Z_2$ and the other supplies keys and values $K = W_K Z_1$, $V = W_V Z_1$:
\begin{align}
\label{eq:cross_attn}
    F &= \Phi_{\mathrm{attn}}(Z_1, Z_2) \nonumber \\
    &= Z_2 + \alpha(Z_2, Z_1)\, W_V Z_1, \nonumber \\
    \alpha(Z_2, Z_1) &:= \mathrm{softmax}\!\left(\frac{(W_Q Z_2)(W_K Z_1)^\top}{\sqrt{d}}\right).
\end{align}

\begin{proposition}[What Architectural Fusion Extracts from $(Z_1, Z_2)$]
\label{prop:fusion_arch}
Let $F = \Phi_{\mathrm{attn}}(Z_1, Z_2)$ be the cross-attention output~\eqref{eq:cross_attn} computed from any local encoders satisfying $Z_i \perp (S, X_{\neq i}, Z_{\neq i}) \mid X_i$. Then:
\begin{enumerate}
    \item \textbf{Upstream information ceiling:} Regardless of the capacity of $\Phi_{\mathrm{attn}}$, fusion cannot recover world-state information discarded by the encoders:
    \begin{equation}
    \label{eq:fusion_dpi}
        I(S; F) \;\le\; I(S; Z_1, Z_2) \;\le\; I(S; X_1, X_2).
    \end{equation}
    In particular, if $(Z_1, Z_2)$ were trained via exact AIB (Corollary~\ref{cor:annihilation}), any fusion module satisfies $I(S; F) \le \Red(S; X_1, X_2)$.
    \item \textbf{Dynamic alignment without global encoder collapse:} Unlike the static alignment objective~\eqref{eq:aib_obj}, cross-attention aligns modalities \emph{conditionally} within the query--key projection $(W_Q Z_2, W_K Z_1)$ while routing uncollapsed values $W_V Z_1$ alongside $Z_2$. Whenever $(Z_1, Z_2)$ satisfies Theorem~\ref{thm:sufficiency}, $F$ can express conditional cross-modal interactions $\Phi_{Z_2}(Z_1)$ so that $I(S; F)$ recovers all four PID atoms ($\Red, \Unq_1, \Unq_2, \Syn$) up to $I(S; Z_1, Z_2)$.
    \item \textbf{Selective architectural bottleneck:} For any subset of modality-1 tokens $B \subset \{1, \dots, N_1\}$ receiving zero attention weight ($\alpha_{\cdot, B} = 0$ a.s.), the fused output satisfies $I(Z_{1,B}; F \mid Z_2, Z_{1, B^c}) = 0$, filtering irrelevant tokens inside the forward pass rather than discarding them at the encoder.
\end{enumerate}
\end{proposition}

The proof is given in Appendix~\ref{app:proofs}. Proposition~\ref{prop:fusion_arch} shows that architectural cross-attention can align modalities toward the world state $S$ without collapsing encoders: alignment occurs dynamically per query in $(W_Q, W_K)$ rather than globally via optimization. However, general fusion over unfactored $(Z_1, Z_2)$ leaves two gaps: it relies entirely on downstream supervision to learn cross-modal correspondence from scratch, and without an explicit pretraining bottleneck it offers no separation between shared redundancy, modality-specific detail, and raw sensor noise. Section~\ref{sec:method} resolves both gaps.

\section{Method}
\label{sec:method}

Theorem~\ref{thm:alignment_redundancy} shows that AIB collapses non-redundant world information, whereas Theorem~\ref{thm:sufficiency} shows that maximizing $I(Z_i; X_i)$ preserves full sufficiency. However, by the chain rule applied to $(Z_i, X_i, S)$, maximizing raw $I(Z_i; X_i)$ unconstrained also amplifies world-irrelevant nuisance: since $I(Z_i; X_i) = \Red(S; Z_1, Z_2) + \Unq(S; Z_i) + I(Z_i; X_i \mid S)$ (derived in Appendix~\ref{app:proofs}), the objective rewards both world-state information and world-irrelevant noise $I(Z_i; X_i \mid S)$. We therefore formulate our method as a \textbf{noise-constrained maximization} of the final encoder decomposition~\eqref{eq:encoder_pid}, realized via subspace decomposition over $(c, u_1, u_2)$ and cross-attention fusion.

\subsection{Constrained Maximization of the Encoder Decomposition}
\label{sec:constrained_max}

Taking the final form of Definition~\ref{def:causal_decomp} in~\eqref{eq:encoder_pid}, we maximize both the modality-unique and cross-modal terms across modalities subject to an explicit cap on world-irrelevant nuisance noise:
\begin{equation}
\label{eq:constrained_opt}
    \max_{q_1, q_2} \;\; \sum_{i=1}^{2} \Big( I(Z_i; X_i \mid X_{\neq i}) + I(Z_i; X_{\neq i}) \Big)
    \qquad \text{subject to} \qquad
    I(Z_i; X_i \mid S) \le \epsilon_i, \quad i \in \{1, 2\},
\end{equation}
where $\epsilon_i \ge 0$ bounds the nuisance capacity retained by encoder $q_i$.

\begin{proposition}[Lagrangian Form and PID Behavior]
\label{prop:constrained_lagrangian}
Introducing Lagrange multipliers $\lambda_i = 1 + \beta > 1$ ($\beta > 0$) for the noise constraints in~\eqref{eq:constrained_opt}, the Lagrangian objective
\begin{equation}
\label{eq:lagrangian_dib}
    \mathcal{J}(Z_1, Z_2)
    \;=\; \sum_{i=1}^{2} \Big( I(Z_i; X_i \mid X_{\neq i}) + I(Z_i; X_{\neq i}) \Big)
    \;-\; (1 + \beta) \sum_{i=1}^{2} I(Z_i; X_i \mid S)
\end{equation}
is identically equal in World-State PID terms to
\begin{equation}
\label{eq:lagrangian_pid}
    \mathcal{J}(Z_1, Z_2)
    \;=\; 2\,\Red(S; Z_1, Z_2) + \sum_{i=1}^{2} \Unq(S; Z_i)
    \;-\; \beta \sum_{i=1}^{2} I(Z_i; X_i \mid S).
\end{equation}
\end{proposition}

The proof is given in Appendix~\ref{app:proofs}. Unlike $\mathcal{J}_{\mathrm{AIB}}$ in~\eqref{eq:aib_pid_exact}, the Lagrangian~\eqref{eq:lagrangian_pid} places strictly positive weight on \textbf{both} $\Red(S; Z_1, Z_2)$ and $\Unq(S; Z_i)$ while placing a strictly negative weight $-\beta < 0$ on noise $I(Z_i; X_i \mid S)$ (requiring $\lambda_i > 1$ because $I(Z_i; X_i)$ itself contains $+1 \cdot I(Z_i; X_i \mid S)$). By Theorem~\ref{thm:sufficiency}, retaining maximal $\Red$ and $\Unq_i$ preserves joint sufficiency so $\Syn(S; Z_1, Z_2)$ remains recoverable.

\subsection{Tractable Training Objective with Noise-Removal Constraints}
\label{sec:tractable_loss}

Because $S$ is unobserved during self-supervised pretraining and exact mutual information is intractable in $\mathbb{R}^k$, we instantiate~\eqref{eq:constrained_opt}--\eqref{eq:lagrangian_dib} through three differentiable terms:

\paragraph{1. Maximizing $I(Z_i; X_i \mid X_{\neq i}) + I(Z_i; X_{\neq i})$ via semantic masked prediction ($\mathcal{L}_{\mathrm{rec}}$).}
Since $I(Z_i; X_i \mid X_{\neq i}) + I(Z_i; X_{\neq i}) = I(Z_i; X_i)$, maximizing intra-modal predictive information recovers both shared and unique structure. To simultaneously enforce the \textbf{semantic noise filter} (removing pixel/sensor noise $I(Z_i; X_i \mid S)$ without observing $S$), we predict abstract target-backbone latents $\bar{h}_i(x_i^{\mathrm{tgt}})$ rather than raw inputs $x_i$:
\begin{equation}
\label{eq:method_rec}
    \mathcal{L}_{\mathrm{rec}}
    \;=\; \sum_{i=1}^{2} \mathbb{E}_{x_i}\!\left[\left\| g_{\psi_i}\!\big(Z_i(x_i^{\mathrm{ctx}})\big) - \operatorname{sg}\!\big(\bar{h}_i(x_i^{\mathrm{tgt}})\big) \right\|_2^2\right],
\end{equation}
where $x_i^{\mathrm{ctx}}$ and $x_i^{\mathrm{tgt}}$ are complementary context/masked views, $g_{\psi_i}$ is a lightweight predictor, and $\operatorname{sg}(\cdot)$ denotes stop-gradient.

\paragraph{2. Reinforcing cross-modal shared information $I(Z_i; X_{\neq i})$ ($\mathcal{L}_{\mathrm{align}}$).}
To explicitly encourage $I(Z_i; X_{\neq i})$, we pair $\mathcal{L}_{\mathrm{rec}}$ with a scale-invariant cosine alignment loss across modalities:
\begin{equation}
\label{eq:method_align}
    \mathcal{L}_{\mathrm{align}}
    \;=\; \mathbb{E}_{(x_1, x_2)}\!\left[\left\| \hat{Z}_1(x_1) - \hat{Z}_2(x_2) \right\|_2^2\right],
    \qquad \hat{Z}_i := \frac{Z_i}{\|Z_i\|_2 + \epsilon}.
\end{equation}

\paragraph{3. Enforcing the noise-capacity constraint $I(Z_i; X_i \mid S) \le \epsilon_i$ ($\mathcal{L}_{\mathrm{sig}}$).}
Even with semantic targets $\bar{h}_i$, unconstrained encoders can collapse or allocate excess capacity to modality-specific noise. We enforce an isotropic Gaussian distribution constraint via empirical characteristic function regularization ($\mathrm{SIGReg}$):
\begin{equation}
\label{eq:method_sig}
    \mathcal{L}_{\mathrm{sig}} \;=\; \sum_{i=1}^{2} \mathcal{L}_{\mathrm{SIGReg}}\big(Z_i(X_i)\big),
\end{equation}
which prevents dimensional collapse while bounding representation channel capacity.

\paragraph{Monolithic baseline objective.}
Combining~\eqref{eq:method_rec}--\eqref{eq:method_sig} on an unsplit latent $Z_i$ yields
\begin{equation}
\label{eq:method_total}
    \mathcal{L} \;=\; \lambda_{\mathrm{rec}}\,\mathcal{L}_{\mathrm{rec}} + \lambda_{\mathrm{align}}\,\mathcal{L}_{\mathrm{align}} + \lambda_{\mathrm{sig}}\,\mathcal{L}_{\mathrm{sig}}.
\end{equation}

\subsection{Subspace Decomposition over Core $c$ and Private Fibers $(u_1, u_2)$}
\label{sec:subspace_decomp}

Optimizing~\eqref{eq:method_total} over a single monolithic vector $Z_i$ forces an inherent tension: $\mathcal{L}_{\mathrm{align}}$ pulls $Z_1$ and $Z_2$ toward a common subspace (which, by Theorem~\ref{thm:alignment_redundancy}, suppresses unique information), whereas $\mathcal{L}_{\mathrm{rec}}$ tries to retain all modality-specific variation inside the same coordinates.

To resolve this conflict directly from~\eqref{eq:encoder_pid} and Proposition~\ref{prop:constrained_lagrangian}, we factorize each encoder representation into a \textbf{shared core} $C_i$ (aligned across modalities into $c$) and a \textbf{private fiber} $U_i$:
\begin{equation}
\label{eq:cu_factor}
    Z_i \;=\; (C_i, U_i), \qquad c_i(x_i) \in \mathbb{R}^{d_c}, \quad u_i(x_i) \in \mathbb{R}^{d_u}, \quad k = d_c + d_u.
\end{equation}
Applying the chain rule to $I(Z_i; X_i)$ yields
\begin{equation}
\label{eq:cu_chain}
    I(Z_i; X_i) \;=\; I(C_i, U_i; X_i) \;=\; \underbrace{I(C_i; X_i)}_{\text{Core } c} \;+\; \underbrace{I(U_i; X_i \mid C_i)}_{\text{Fiber } u_i},
\end{equation}
which mirrors the two terms of~\eqref{eq:encoder_pid}: $C_i$ is dedicated to the cross-modal term $I(Z_i; X_{\neq i})$ ($\Red$), while $U_i$ is dedicated to the conditional residual $I(Z_i; X_i \mid X_{\neq i})$ ($\Unq_i$). We apply the losses from Section~\ref{sec:tractable_loss} in a decomposed manner to each latent subspace:

\paragraph{1. Shared core loss ($\mathcal{L}_c$).}
The core coordinates $(c_1, c_2)$ receive only the scale-invariant cosine alignment loss~\eqref{eq:method_align} and an isolated $\mathrm{SIGReg}$ bottleneck~\eqref{eq:method_sig}:
\begin{equation}
\label{eq:loss_c}
    \mathcal{L}_{c}
    \;=\; \lambda_{\mathrm{align}}\,\mathbb{E}_{(x_1, x_2)}\!\left[\left\| \hat{c}_1(x_1) - \hat{c}_2(x_2) \right\|_2^2\right]
    \;+\; \lambda_{\mathrm{sig}} \sum_{i=1}^{2} \mathcal{L}_{\mathrm{SIGReg}}\big(c_i(X_i)\big),
    \qquad \hat{c}_i := \frac{c_i}{\|c_i\|_2 + \epsilon}.
\end{equation}
By Theorem~\ref{thm:alignment_redundancy}, restricting alignment to $c_1, c_2$ extracts $\Red(S; X_1, X_2)$ into the shared core $c$ without collapsing modality-unique information. Unit normalization prevents centroid magnitude collapse under one-to-many asymmetry, while applying $\mathcal{L}_{\mathrm{SIGReg}}$ to $c_i$ separately prevents $u_i$ from absorbing all variance and leaving $c_i$ degenerate.

\paragraph{2. Private fiber losses ($\mathcal{L}_{u_1}, \mathcal{L}_{u_2}$).}
Each modality's private fiber $u_i$ receives the intra-modal masked JEPA prediction loss~\eqref{eq:method_rec} conditioned on the stop-gradiented core (reflecting $I(U_i; X_i \mid C_i)$ in~\eqref{eq:cu_chain}), paired with its own $\mathrm{SIGReg}$ capacity constraint:
\begin{align}
\label{eq:loss_ui}
    \mathcal{L}_{u_i} &= \lambda_{\mathrm{rec}}\,\mathbb{E}_{x_i}\!\left[\left\| g_{\psi_i}\!\big(u_i(x_i^{\mathrm{ctx}}),\, \operatorname{sg}(c_i(x_i^{\mathrm{ctx}}))\big) - \operatorname{sg}\!\big(\bar{h}_i(x_i^{\mathrm{tgt}})\big) \right\|_2^2\right] \nonumber \\
    &\quad + \lambda_{\mathrm{sig}}\,\mathcal{L}_{\mathrm{SIGReg}}\big(u_i(X_i)\big), \qquad i \in \{1, 2\}.
\end{align}
Conditioning $g_{\psi_i}$ on $\operatorname{sg}(c_i)$ treats the aligned core as given context so $u_i$ encodes strictly residual modality-unique structure ($\Unq_i$), while target-latent prediction and $\mathcal{L}_{\mathrm{SIGReg}}(u_i)$ suppress raw sensor noise.

\paragraph{3. Within-modality core--fiber independence ($\mathcal{L}_{\mathrm{ind}}$).}
To prevent shared semantics and private residuals from duplicating each other across subspaces within the same modality, we penalize the batch-normalized cross-covariance strictly between $u_i$ and $c_i$:
\begin{equation}
\label{eq:loss_ind}
    \mathcal{L}_{\mathrm{ind}}
    \;=\; \sum_{i=1}^{2} \left\| \widehat{\Cov}\big(u_i(X_i),\, c_i(X_i)\big) \right\|_F^2.
\end{equation}
Crucially, no independence penalty is imposed between $u_1$ and $u_2$, allowing cross-modal coordination to remain intact for downstream synergy.

\paragraph{Total decomposed objective, fusion, and why decomposition is superior.}
Combining~\eqref{eq:loss_c}--\eqref{eq:loss_ind}, the decomposed pretraining loss is
\begin{equation}
\label{eq:loss_djepa_total}
    \mathcal{L}_{\mathrm{D\text{-}JEPA}}
    \;=\; \mathcal{L}_{c} + \mathcal{L}_{u_1} + \mathcal{L}_{u_2} + \lambda_{\mathrm{ind}}\,\mathcal{L}_{\mathrm{ind}}.
\end{equation}
For any downstream task $T = h(S)$, the predictors $g_{\psi_i}$ and target backbones $\bar{h}_i$ are discarded, and a cross-attention fusion head $\Phi_{\mathrm{attn}}(c, u_1, u_2)$~\eqref{eq:cross_attn} operates over the decomposed bundle (Table~\ref{tab:subspace_losses}). Combined with Proposition~\ref{prop:fusion_arch}, D-JEPA strictly improves over all three alternatives:
\begin{itemize}
    \item \textbf{Over alignment-only:} Unlike AIB (Theorem~\ref{thm:alignment_redundancy}), $u_1$ and $u_2$ satisfy Theorem~\ref{thm:sufficiency} on modality-unique residuals, lifting the upstream ceiling~\eqref{eq:fusion_dpi} from $\Red$ to full $I(S; X_1, X_2)$ so $\Phi_{\mathrm{attn}}$ can extract $\Syn$.
    \item \textbf{Over monolithic hybrids:} Unlike~\eqref{eq:method_total}, isolating alignment in $c$ eliminates coordinate conflict between $\mathcal{L}_{\mathrm{align}}$ and $\mathcal{L}_{\mathrm{rec}}$.
    \item \textbf{Over unaligned infomax + fusion:} Unlike raw $(Z_1, Z_2)$ fed to $\Phi_{\mathrm{attn}}$, D-JEPA pre-aligns shared semantics in $c$, filters sensor noise via $\mathcal{L}_{\mathrm{SIGReg}}$ and semantic JEPA targets, and provides a single-modality fallback floor whenever modality $\neq i$ is missing:
    \begin{equation}
    \label{eq:dropout_floor}
        I(S; C_i, U_i) \;=\; I(S; X_i) - I(S; X_i \mid C_i, U_i) \;=\; \Red(S; X_1, X_2) + \Unq(S; X_i) - \delta_i,
    \end{equation}
    where $\delta_i := I(S; X_i \mid C_i, U_i)$ is the marginal residual gap.
\end{itemize}

\section{Experiments}
\label{sec:experiments}

\subsection{Experimental Protocol}
We evaluate across two complementary suites designed to isolate the mechanics of representation learning under asymmetry:
\begin{itemize}
    \item \textbf{Synthetic Controlled Ground-Truth Benchmark}: To evaluate our theorems where
    ground-truth PID atoms are mathematically known, we generate paired streams $(X_1, X_2)$
    from non-linear mixtures of four independent latent factors: a shared factor $S \in \mathbb{R}^{d}$,
    modality-private factors $U_1, U_2 \in \mathbb{R}^{d}$, and a non-linear synergistic parity
    factor $Y_{\mathrm{syn}} \in \mathbb{R}^{d}$ ($d=64$, $N=3{,}000$ paired samples per seed across 2 seeds).
    \item \textbf{Real Multimodal Benchmark (CLEVR)}: To test the architecture on real-world
    informational asymmetry where questions act as selective relational operators over dense scenes,
    we train on $12{,}000$ image--question pairs ($3{,}000$ Blender-rendered CLEVR scenes with 6 relational
    questions each), evaluated across 3 random seeds. Questions are categorized into functional types:
    \emph{Attribute}, \emph{Counting}, \emph{Exist/Compare}, and \emph{Other}. Encoders use frozen DINOv2
    ViT-S/14 ($4 \times 4$ patch grid) and BERT-base ($T=32$).
    \item \textbf{Fair Baseline Arms}: Every model arm uses identical encoder architectures,
    parameter counts, and total latent dimensionality ($4 \times d_{\mathrm{comp}} = 256$):
    (1) \textbf{Raw}: Concatenated frozen features ($h_1 \,\|\, h_2$);
    (2) \textbf{Alignment-only}: Standard cross-modal invariance and SIGReg;
    (3) \textbf{Alignment-only + Rec}: Alignment-only augmented with contextual predictors;
    (4) \textbf{D-JEPA (No-Indep)}: D-JEPA setting $w_{\mathrm{ind}} = 0$;
    (5) \textbf{D-JEPA (No-Rec)}: D-JEPA setting $w_{\mathrm{rec}} = 0$;
    (6) \textbf{D-JEPA (Full Proposed)}: The complete factorized architecture.
\end{itemize}

\subsection{Downstream Reasoning and Modality Gap Recovery}

\begin{table}[t]
\centering
\small
\caption{\textbf{Downstream representation probing on the synthetic benchmark}
(test set, mean $\pm$ std). Alignment-only training collapses on private and synergistic factors ($48.33\%$ on synergy),
producing an inverted synergy gap ($-3.20\,\text{pp}$). D-JEPA preserves factor decodability across all atoms ($84.20\%$ on synergy),
restoring a positive multimodal synergy gap ($+33.13\,\text{pp}$) matching the raw ceiling ($+35.00\,\text{pp}$). The
$\Delta_{\mathrm{joint}-\max(v,t)}$ column is computed as $\mathrm{acc}(\text{joint}) - \max(\mathrm{acc}(\text{vision}), \mathrm{acc}(\text{text}))$.}
\label{tab:main_results}
\resizebox{\textwidth}{!}{%
\begin{tabular}{l cccc c}
\toprule
\textbf{Model / Arm} & \textbf{Shared (\%)} & \textbf{Private Vis (\%)} & \textbf{Private Txt (\%)} & \textbf{Synergy (\%)} & \textbf{Synergy Gap $\Delta_{\mathrm{joint}-\max(v,t)}$ (pp)} \\
\midrule
Raw (Frozen Backbone Ceiling) & $84.60 \pm 3.11$ & $79.47 \pm 1.13$ & $77.33 \pm 1.70$ & $85.87 \pm 0.19$ & $+35.00 \pm 0.09$ \\
\midrule
Alignment-only $[z_1, z_2]$     & $81.07 \pm 2.07$ & $55.33 \pm 3.77$ & $51.93 \pm 10.65$ & $48.33 \pm 2.73$ & $-3.20 \pm 2.83$ \\
Alignment-only + Rec           & $81.27 \pm 2.17$ & $69.47 \pm 1.13$ & $67.80 \pm 2.55$ & $49.73 \pm 0.94$ & $-1.67 \pm 2.36$ \\
\midrule
\textbf{D-JEPA (No-Independence)} & $\mathbf{83.00 \pm 1.79}$ & $\mathbf{77.60 \pm 0.75}$ & $\mathbf{75.40 \pm 3.68}$ & $\mathbf{86.87 \pm 1.04}$ & $\mathbf{+35.73 \pm 1.70}$ \\
D-JEPA (No-Reconstruction)      & $78.27 \pm 2.83$ & $75.27 \pm 1.98$ & $72.87 \pm 1.04$ & $83.00 \pm 0.85$ & $+32.13 \pm 1.51$ \\
D-JEPA (Full Proposed) & $79.40 \pm 2.36$ & $74.87 \pm 1.41$ & $73.80 \pm 1.04$ & $84.20 \pm 0.47$ & $+33.13 \pm 0.47$ \\
\bottomrule
\end{tabular}%
}
\end{table}

\begin{table}[t]
\centering
\small
\caption{\textbf{Downstream compositional reasoning on CLEVR} (3 seeds, mean $\pm$ std, \%).
Alignment-only training collapses on fine-grained visual reasoning categories (Attribute: $0.43\%$, Counting: $4.88\%$), surviving only on binary queries (Exist: $48.90\%$), which inverts the synergy gap to $-1.56\,\text{pp}$. D-JEPA prevents dimensional collapse, restoring positive multimodal synergy ($+0.54\,\text{pp}$) and gaining $\mathbf{+18.36\,\text{pp}}$ overall ($95\%$ paired bootstrap CI $[17.21, 19.44]\,\text{pp}$). In every row, the text probe exceeds the vision probe, so $\Delta_{\mathrm{joint}-\max(v,t)} = \Delta_{\mathrm{joint}-\text{text}}$.}
\label{tab:clevr_main}
\resizebox{\textwidth}{!}{%
\begin{tabular}{l ccc c cccc c}
\toprule
 & \multicolumn{3}{c}{\textbf{Modality Probing Acc (\%)}} & \textbf{Synergy Gap} & \multicolumn{4}{c}{\textbf{Category Breakdown (\%)}} & \textbf{Paired Gain vs.} \\
\cmidrule(lr){2-4} \cmidrule(lr){5-5} \cmidrule(lr){6-9}
\textbf{Model / Arm} & \textbf{Vision ($v$)} & \textbf{Text ($t$)} & \textbf{Joint ($v+t$)} & $\Delta_{\mathrm{joint}-\max(v,t)}$ & \textbf{Attribute} & \textbf{Counting} & \textbf{Exist/Cmp} & \textbf{Other} & \textbf{Alignment-only (CI)} \\
\midrule
Raw (Frozen Ceiling) & $20.79 \pm 0.21$ & $42.02 \pm 0.79$ & $45.57 \pm 0.35$ & $+3.55 \pm 0.47$ & $42.83 \pm 0.25$ & $37.08 \pm 0.44$ & $52.49 \pm 0.85$ & $37.96 \pm 0.37$ & $+24.32\ [23.18, 25.46]$ \\
\midrule
Alignment-only $[z_1, z_2]$ & $20.50 \pm 0.42$ & $22.81 \pm 1.11$ & $21.25 \pm 0.69$ & $-1.56 \pm 0.65$ & $0.43 \pm 0.64$ & $4.88 \pm 3.93$ & $48.90 \pm 2.72$ & $4.80 \pm 4.82$ & --- \\
Alignment-only + Rec & $21.33 \pm 0.20$ & $29.14 \pm 2.18$ & $27.96 \pm 2.22$ & $-1.19 \pm 0.11$ & $10.61 \pm 0.98$ & $24.35 \pm 3.95$ & $45.59 \pm 2.33$ & $23.55 \pm 4.47$ & $+6.70\ [5.75, 7.59]^*$ \\
\midrule
D-JEPA (No-Indep) & $21.25 \pm 0.12$ & $28.44 \pm 1.10$ & $27.99 \pm 1.54$ & $-0.45 \pm 0.81$ & $11.10 \pm 1.98$ & $24.35 \pm 1.70$ & $44.37 \pm 1.51$ & $26.39 \pm 3.16$ & $+6.74\ [5.82, 7.66]^*$ \\
D-JEPA (No-Rec) & $20.49 \pm 0.40$ & $36.39 \pm 5.90$ & $35.38 \pm 6.40$ & $-1.01 \pm 0.52$ & $25.92 \pm 9.81$ & $29.30 \pm 2.99$ & $47.65 \pm 4.38$ & $27.70 \pm 6.42$ & $+14.13\ [12.89, 15.37]^*$ \\
\textbf{D-JEPA (Full Proposed)} & $\mathbf{20.96 \pm 0.29}$ & $\mathbf{39.07 \pm 2.46}$ & $\mathbf{39.61 \pm 3.07}$ & $\mathbf{+0.54 \pm 0.65}$ & $\mathbf{29.69 \pm 8.01}$ & $\mathbf{32.41 \pm 1.25}$ & $\mathbf{51.80 \pm 0.62}$ & $\mathbf{34.46 \pm 1.18}$ & $\mathbf{+18.36\ [17.21, 19.44]^*}$ \\
\bottomrule
\end{tabular}%
}
\end{table}

The evaluations in Tables~\ref{tab:main_results} and~\ref{tab:clevr_main} confirm our theoretical predictions:

\begin{enumerate}
    \item \textbf{Collapse of alignment-only baselines}: In synthetic benchmarks
    (Table~\ref{tab:main_results}), alignment-only training retains shared structure ($81.07\%$)
    but collapses to chance on synergistic factors ($48.33\%$). On real CLEVR data
    (Table~\ref{tab:clevr_main}), this manifests as relational collapse: overall accuracy
    drops to $21.25 \pm 0.69\%$, falling to $0.43\%$ on attributes and $4.88\%$ on counting.
    Only binary queries (exist/compare, $48.90\%$) survive via majority-class guessing.
    Adding reconstruction without latent factorization (Alignment-only + Rec) improves overall
    accuracy to $27.96\%$, but remains severely depressed.
    \item \textbf{Multimodal Synergy Gap Inversion}: Under alignment-only training, adding vision
    representations to language probes \emph{actively degrades} performance compared to text probes alone,
    inverting the synergy gap to negative territory ($-3.20\,\text{pp}$ on synthetic, $-1.56\,\text{pp}$ on CLEVR).
    D-JEPA resolves this pathology on the synthetic benchmark, recovering positive multimodal synergy
    ($+33.13 \pm 0.47\,\text{pp}$), while on CLEVR it converts the alignment-only inversion into a
    non-significantly-negative gap ($+0.54 \pm 0.65\,\text{pp}$ vs.\ alignment-only's $-1.56 \pm 0.65\,\text{pp}$).
    \item \textbf{D-JEPA Restores Compositional Factorization}: D-JEPA achieves $\mathbf{39.61 \pm 3.07\%}$
    overall accuracy on CLEVR ($\mathbf{+18.36\,\text{pp}}$ gain over alignment-only, $95\%$ bootstrap CI
    $[17.21, 19.44]\,\text{pp}$), capturing $87\%$ of the frozen backbone ceiling ($45.57\%$). Performance
    on fine-grained attributes rebounds from $0.43\%$ to $29.69\%$, and counting rebounds from $4.88\%$ to $32.41\%$.
    Notably, on the synthetic benchmark (Table~\ref{tab:main_results}), D-JEPA's full-latent probe on the
    shared factor ($79.40\%$) is slightly below the alignment-only baseline ($81.07\%$): a $1.67\,\text{pp}$
    reduction in shared decodability, well within the reported std. D-JEPA trades a small amount of shared
    decodability for large gains on private and synergistic structure, which is the empirically observed form
    of the escape from the bound.
\end{enumerate}

\paragraph{CLEVR as a downstream demonstration, not a test of the bound.}
On CLEVR, linear cross-modal CKA for the alignment-only arm is $0.016$ against a
shuffled-pair null of $0.004$ (Table~\ref{tab:alignment}), two orders of
magnitude below the synthetic benchmark's $0.914$: under this degree of
asymmetry there is no common linear view for the encoders to converge to. The
alignment-only model's downstream collapse ($21.25\%$ against the $45.57\%$
frozen ceiling) is therefore not a measurement of
Theorem~\ref{thm:alignment_redundancy}; it is a measurement of what happens when
an alignment objective is applied to pairs that do not satisfy its premise.
D-JEPA's shared code is also at the null level on CLEVR ($0.001$ vs $0.001$),
which we read as further evidence that CLEVR is a regime in which no linear
shared core exists: the recovery ($+18.36$~pp) is attributable to the private
fibers and the JEPA grounding, not to recovering an aligned core. The theorem is
tested on the synthetic benchmark, where alignment is achieved and the PID atoms
are known; CLEVR demonstrates the downstream consequence of applying the
objective to data it was not designed for.

\paragraph{Ablation interaction with data regime.}
An apparent tension in Table~\ref{tab:main_results} warrants direct comment:
on the synthetic benchmark, D-JEPA (No-Independence) matches or exceeds the
Full model across all four factor categories ($83.00$, $77.60$, $75.40$,
$86.87$ vs.\ $79.40$, $74.87$, $73.80$, $84.20$). We read this as evidence that
\textbf{factorization with a private fiber is the load-bearing ingredient, and
the independence penalty is a tuning knob rather than a required component
when the generative process already separates shared and private factors}.
On CLEVR, where real frozen features entangle shared and private content, the
regime reverses: the Full model outperforms No-Independence by
$11.62\,\text{pp}$ ($39.61\%$ vs.\ $27.99\%$, Table~\ref{tab:clevr_main}),
indicating that the independence penalty becomes essential when the underlying
features are not already separated by construction. We retain
$\mathcal{L}_{\mathrm{ind}}$ as the default because the real-data regime is
where factorization is most needed; we do not claim the penalty is required
in the idealized synthetic regime. The causal fiber-swap intervention
(\S\ref{sec:swap}) provides independent evidence that the recovered
information is carried by the private fiber rather than by the specific loss
combination: swapping $u_2$ flips $71.36\%$ of predictions on CLEVR, confirming
causal binding.

\subsection{Factor Dissection: What Lives in Each Subspace}
\label{sec:factor_dissection}

To verify that D-JEPA's learned subspaces achieve the designed functional decomposition, we perform linear regression audits ($R^2$; Table~\ref{tab:clevr_subspaces} and Appendix~\ref{app:idmat}).

\begin{table}[t]
\centering
\small
\caption{\textbf{CLEVR subspace specialization and factor decodability.}
(a)~Probe accuracy from isolated D-JEPA subspace blocks on downstream CLEVR answers.
(b)~Linear regression decodability ($R^2$) of question type (\texttt{qtype}) and mentioned entities (\texttt{mentions}). Alignment-only collapses question semantics ($R^2 = 0.159$), whereas D-JEPA retains robust linguistic structure across $c_2$ and $u_2$ ($R^2 = 0.713$).}
\label{tab:clevr_subspaces}

\resizebox{\textwidth}{!}{%
\begin{tabular}{l ccccc}
\toprule
\multicolumn{6}{l}{\textbf{(a) Isolated D-JEPA Subspace Probing Accuracy on CLEVR Answers (\%)}} \\
\midrule
\textbf{Metric} & \textbf{Shared Pair ($c$)} & \textbf{Private Pair ($u$)} & \textbf{Vision Private ($u_1$)} & \textbf{Text Shared ($c_2$)} & \textbf{Text Private ($u_2$)} \\
\midrule
Probe Accuracy & $35.76 \pm 2.57$ & $32.07 \pm 0.59$ & $20.99 \pm 0.21$ & $36.62 \pm 2.58$ & $34.65 \pm 1.82$ \\
\bottomrule
\end{tabular}%
}

\vspace{0.7em}

\resizebox{\textwidth}{!}{%
\begin{tabular}{l ccc l}
\toprule
\multicolumn{5}{l}{\textbf{(b) Factor Identification Matrix ($R^2$ Held-Out Decodability)}} \\
\midrule
\textbf{Subspace Stream} & \textbf{Question Type (\texttt{qtype})} & \textbf{Mentioned Concepts (\texttt{mentions})} & \textbf{Total Stream $R^2$} & \textbf{Functional Role / Interpretation} \\
\midrule
Raw Language Backbone ($h_2$)      & $0.963$ & $0.844$ & ---     & Frozen linguistic ceiling \\
Alignment-only Text Latent ($z_2$)  & $0.159$ & $0.080$ & $0.159$ & \textbf{Severe semantic collapse} \\
Alignment-only + Rec Text ($z_2$)  & $0.452$ & $0.184$ & $0.452$ & Partial reconstruction rescue \\
\midrule
D-JEPA Shared Text ($c_2$)         & $0.499$ & $0.332$ & \multirow{2}{*}{$\mathbf{0.713}$} & Grounded question component \\
D-JEPA Private Text ($u_2$)        & $0.583$ & $0.318$ &         & Private relational query operator \\
\bottomrule
\end{tabular}%
}
\end{table}

Under Alignment-only training on CLEVR, forcing language representations to predict visual features pulls $z_2$ toward generic visual centroids: question type decodability collapses from $R^2 = 0.963$ to $0.159$ (an $83\%$ loss), and entity referents drop from $0.844$ to $0.080$ (a $90\%$ loss). In D-JEPA, question semantics are preserved across the factorized text stream ($R^2 = 0.713$), with private fiber $u_2$ capturing the unaligned relational logic ($R^2 = 0.583$) and shared code $c_2$ capturing grounded concepts ($R^2 = 0.499$). Isolated probing confirms that $c$ alone reaches $35.76\%$ and $u$ reaches $32.07\%$, verifying that both streams carry distinct task-relevant information.

\subsection{Representation Geometry and Latent Disentanglement}
\label{sec:geometry}

\begin{figure}[t]
\centering
\includegraphics[width=0.72\textwidth]{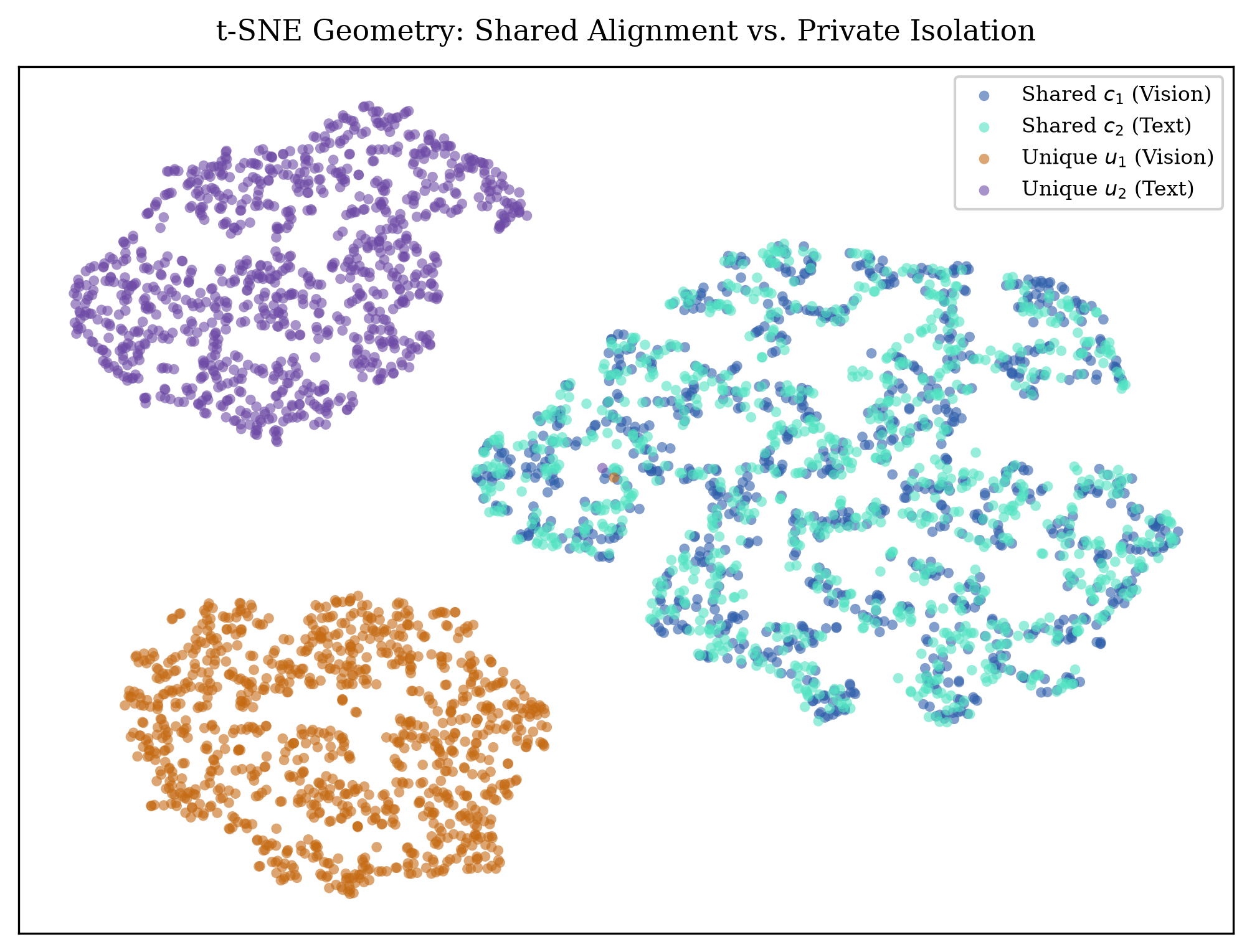}
\caption{\textbf{t-SNE latent geometry of D-JEPA subspaces.} Held-out latents of the four learned codes: shared $c_1$ (vision, blue) and $c_2$ (text, teal) exhibit close cross-modal alignment, while private fibers $u_1$ (vision, orange) and $u_2$ (text, purple) occupy isolated, non-overlapping subspaces.}
\label{fig:tsne}
\end{figure}

Figure~\ref{fig:tsne} visualizes the held-out latent space geometry via t-SNE. Under Alignment-only training, representations collapse onto a unified shared manifold, reflecting excessive invariance. In D-JEPA, the four distinct codes organize into structured, well-separated geometric manifolds: shared codes $c_1$ (vision) and $c_2$ (text) align in representation space, while private fibers $u_1$ (vision) and $u_2$ (text) occupy isolated, non-overlapping subspaces.

\subsection{Information-Theoretic Audits: Probe-PID and Flow-PID}
\label{sec:pid_audit}

We audit representations using discrete Probe-PID and continuous Flow-PID estimators (Table~\ref{tab:clevr_pid} and Appendix~\ref{app:synthetic_pid}).

\begin{table}[t]
\centering
\small
\caption{\textbf{Information-theoretic audits on CLEVR} (bits).
(Top) Probe-PID on CLEVR answer labels ($I(Y; S_1, S_2)$). Alignment-only destroys mutual information down to $0.242$ bits, whereas D-JEPA restores $1.938$ bits ($84.5\%$ of raw ceiling) and recovers counting synergy. (Bottom) Flow-PID continuous latent density estimation on object count ($I_{12}$).}
\label{tab:clevr_pid}
\resizebox{\textwidth}{!}{%
\begin{tabular}{l ccccc c}
\toprule
\textbf{Model / Stream} & \textbf{Redundancy ($\Red$)} & \textbf{Vision-Unq ($\Unq_1$)} & \textbf{Text-Unq ($\Unq_2$)} & \textbf{Synergy ($\Syn$)} & \textbf{Total $I(Y; S_1, S_2)$} & \textbf{\% Raw Ceiling} \\
\midrule
\multicolumn{7}{l}{\textbf{Probe-PID: Downstream CLEVR Answer Prediction (All Categories)}} \\
Raw ($h_1 \,\|\, h_2$) & $0.010$ & $0.000$ & $2.250$ & $\mathbf{0.033}$ & $\mathbf{2.293}$ bits & $100.0\%$ \\
Alignment-only ($z_1 \,\|\, z_2$) & $0.004$ & $0.000$ & $0.237$ & $0.000$ & $0.242$ bits & $10.6\%$ \\
Alignment-only + Rec & $0.006$ & $0.000$ & $1.000$ & $0.000$ & $1.006$ bits & $43.9\%$ \\
\textbf{D-JEPA (Full Stream)} & $\mathbf{0.002}$ & $\mathbf{0.000}$ & $\mathbf{1.935}$ & $0.000$ & $\mathbf{1.938}$ bits & $\mathbf{84.5\%}$ \\
\quad \emph{Category: Counting Only} & $0.071$ & $0.000$ & $2.075$ & $\mathbf{0.037}$ & $\mathbf{2.182}$ bits & (Raw $\Syn = 0.082$) \\
\quad \emph{Category: Attribute Only} & $0.024$ & $0.000$ & $2.524$ & $0.000$ & $2.548$ bits & (Raw $I = 3.335$) \\
\midrule
\multicolumn{7}{l}{\textbf{Flow-PID: Continuous Latent Density Decomposition (Object Count Target)}} \\
\textbf{Model / Stream} & $\mathbf{I_1}$ & $\mathbf{I_2}$ & $\mathbf{I_{12}}$ & $\mathbf{\Red}$ & $\mathbf{\Syn}$ & $\mathbf{I_{12}^{\mathrm{null}}}$ \\
Raw ($h_1 \,\|\, h_2$) & $0.124$ & $0.348$ & $0.492$ & $0.124$ & $0.143$ & $0.212$ \\
Alignment-only ($z_1 \,\|\, z_2$) & $0.112$ & $0.125$ & $0.249$ & $0.108$ & $0.120$ & $0.211$ \\
\textbf{D-JEPA (Full Stream)} & $0.134$ & $0.225$ & $\mathbf{0.370}$ & $0.134$ & $\mathbf{0.145}$ & $0.219$ \\
\bottomrule
\end{tabular}%
}
\end{table}

The Probe-PID audit on CLEVR (Table~\ref{tab:clevr_pid}) empirically validates the Alignment Redundancy Bound:
alignment-only training obliterates downstream task mutual information, falling from $2.293$ bits (raw)
to $0.242$ bits. D-JEPA restores mutual information to $\mathbf{1.938}$ bits ($84.5\%$ of the ceiling).
On counting queries, where answer determination requires combining visual object features with textual counting criteria,
alignment-only yields $0.000$ bits of synergy, while D-JEPA recovers $\mathbf{0.037}$ bits of pure synergy.
The Flow-PID continuous latent density audit corroborates this finding: D-JEPA maintains
$I_{12} = 0.370$ bits on object count density, comfortably above the null floor of $0.219$ bits,
while alignment-only degrades to $0.249$ bits, only marginally above the null floor of $0.219$.
Probe-PID and Flow-PID target different quantities---discrete answer labels
versus a continuous object-count density---so they agree in direction but not
magnitude; we treat the discrete Probe-PID as the operational test of the bound
and the continuous Flow-PID as a corroborating check that the effect survives a
different estimator.

\subsection{Causal Fiber-Swap Intervention}
\label{sec:swap}

To establish that the learned fibers causally control predictions rather than acting as passive statistical decorrelators, we perform a randomized donor fiber-swap intervention: for each test sample, a single fiber block is replaced by the corresponding block from an independent donor pair (Table~\ref{tab:clevr_swap} and Appendix~\ref{app:synthetic_swap}).

\begin{table}[h]
\centering
\small
\caption{\textbf{Causal donor fiber-swap intervention on CLEVR} (prediction flip rate, mean $\pm$ std, \%).
Replacing a test sample's fiber with that of a random donor demonstrates active causal binding: swapping $u_2$ (private text) flips $71.36\%$ of predictions, while swapping $c_2$ flips $65.82\%$. The private visual fiber $u_1$ causes $35.03\%$ flips, confirming that downstream reasoning actively depends on both private channels.}
\label{tab:clevr_swap}
\resizebox{\textwidth}{!}{%
\begin{tabular}{l cccc}
\toprule
\textbf{Intervention} & \textbf{Swap Shared Vis ($c_1$)} & \textbf{Swap Private Vis ($u_1$)} & \textbf{Swap Shared Txt ($c_2$)} & \textbf{Swap Private Txt ($u_2$)} \\
\midrule
Alignment-only ($z_1$ / $z_2$) & $44.48 \pm 16.78$ & --- & $44.60 \pm 18.33$ & --- \\
Alignment-only + Rec & $52.97 \pm 10.24$ & --- & $81.81 \pm 25.01$ & --- \\
\midrule
D-JEPA (No-Indep) & $13.53 \pm 9.75$ & $42.90 \pm 9.55$ & $24.87 \pm 10.54$ & $77.80 \pm 23.14$ \\
D-JEPA (No-Rec) & $34.93 \pm 7.84$ & $32.80 \pm 8.05$ & $57.51 \pm 16.16$ & $70.01 \pm 19.27$ \\
\textbf{D-JEPA (Full Proposed)} & $\mathbf{27.96 \pm 7.10}$ & $\mathbf{35.03 \pm 7.36}$ & $\mathbf{65.82 \pm 17.17}$ & $\mathbf{71.36 \pm 19.59}$ \\
\bottomrule
\end{tabular}%
}
\end{table}

On CLEVR (Table~\ref{tab:clevr_swap}), swapping $u_2$ flips $71.36 \pm 19.59\%$ of answer predictions, while swapping $c_2$ flips $65.82 \pm 17.17\%$. Swapping the private visual fiber $u_1$ causes $35.03 \pm 7.36\%$ flips, demonstrating that downstream decisions causally bind private perceptual details from vision with relational logic from text.

\section{Related Work}
\label{sec:related}

\paragraph{Contrastive Cross-Modal Alignment: CLIP, ALIGN, and Their Blind Spots.}
The dominant recipe for multimodal representation learning aligns two encoders
through a contrastive objective on paired data. CLIP~\citep{radford2021clip}
and ALIGN~\citep{jia2021align} established that a symmetric InfoNCE loss over
large-scale noisy image--text pairs yields embeddings with strong zero-shot and
retrieval performance, and the CLIP-style dual-encoder has become the default
backbone for cross-modal alignment. The objective itself, however, enforces
invariance rather than preservation: it pulls paired embeddings together and
pushes unpaired embeddings apart, without any term that assigns value to
information present in only one modality.

\paragraph{What alignment discards: gaps and bottlenecks.}
The observation that alignment can discard modality-specific information has
been documented across several literatures. \citet{liang2022mind} identified
the modality gap as a structural property of contrastive training, and
\citet{tsai2019learning} linked it to the loss of view-specific content in
factorized objectives. The information-bottleneck perspective gives a general
account: alignment is a rate constraint, and anything outside the constraint is
discarded. What has been missing is a tight statement of \emph{which}
information is discarded by the specific cross-modal Dirichlet energy that
JEPAs and CLIP-style models minimize. Theorem~\ref{thm:alignment_redundancy}
supplies this: retained information is bounded by the redundancy atom of the
PID decomposition, a distribution-free statement about the alignment term alone. The
bound is a statement about the optimum of a specific objective, not a claim
about what any convergent algorithm must recover; algorithm-level identifiability
is a separate question that we leave open.

\paragraph{Shared-private decomposition.}
Splitting a latent into shared and private parts has a long history in
multi-view learning. Generative approaches such as Multimodal
VAEs~\citep{shi2019variational} use
modality-specific decoders; canonical-correlation approaches such as Deep
CCA~\citep{andrew2013deep} and its extensions learn maximally correlated
subspaces; non-contrastive regularizers such as VICReg~\citep{bardes2022vicreg}
and Barlow Twins~\citep{zbontar2021barlow} prevent collapse through variance and
decorrelation terms without partitioning the latent. D-JEPA differs from each in
specific ways. It uses masked prediction rather than pixel reconstruction to
anchor the private fibers, tying them to task-relevant structure rather than
fine-grained detail. It applies the independence penalty asymmetrically ---
between private and shared, but not between the two private fibers --- which
allows the visual and textual private fibers to coordinate on compositional
binding, the exact capability that pure independence would suppress. And it
evaluates the private fibers through causal and information-theoretic audits,
not just downstream accuracy. Our contribution is not the factorization itself
but the masked-prediction anchor and the asymmetric independence penalty that
let the factorization carry task-relevant information under informational
asymmetry.

\paragraph{Partial Information Decomposition and diagnostic audits.}
PID decomposes the information that multiple sources carry about a target into
redundant, unique, and synergistic
atoms~\citep{williams2010nonnegative,bertschinger2014quantifying}, and has been
used to analyze information sharing in static and dynamical Gaussian
systems~\citep{harder2013bivariate}. Exact PID is
intractable for high-dimensional continuous variables, but recent work has made
it practical through Gaussian and
flow-based~\citep{zhang2025flowpid} estimators. Our Probe-PID and Flow-PID
audits use these estimators as \emph{diagnostics of the training objective}: to
test whether the objective recovers the atoms it is designed to recover. The
framework in Section~\ref{sec:theory} makes a specific prediction --- that
alignment-only representations have near-zero unique and synergistic atoms ---
and the audits in Section~\ref{sec:pid_audit} test it. To our knowledge this is
the first use of PID to test a theoretical claim about representation learning
rather than to measure information sharing in a fixed model.

\paragraph{Compositional reasoning.}
Benchmarks such as CLEVR~\citep{johnson2017clevr} and
Winoground~\citep{thrush2022winoground} have established that models optimizing
for cross-modal agreement fail on compositional queries --- relational,
counting, and binding questions that require connecting a linguistic operator
to specific visual referents. The failure is well-documented but the explanation
is not. Our work proposes a specific explanation and tests it: alignment
discards the binding structure that compositional queries require, and
decomposition recovers it. Section~\ref{sec:experiments} makes this concrete by
isolating the failure to attribute and counting questions, and by showing that
private fibers restore accuracy on exactly those categories.

\section{Discussion, Limitations, and Future Directions}
\label{sec:discussion}

\paragraph{The Limits of Pure Alignment.} Cross-modal alignment is an information projection, not an information-preserving operation. Under cross-modal Dirichlet energy minimization, features absent from both modalities are treated as noise. In asymmetric multimodal settings---where language interrogates a scene and vision provides fine-grained spatial context---aligned models undergo category collapse ($0.43\%$ on attributes, $4.88\%$ on counting). D-JEPA demonstrates that factorizing latents into shared and private subspaces recovers this lost structure without requiring generative pixel-level decoders.

\paragraph{Functional Modularity and Non-Interfering Representations.}
A key architectural insight of D-JEPA is that representation learning benefits from explicit separation between common and private evidence streams. In standard monolithic alignment ($z_1 \approx z_2$), the model forces the language encoder to predict visual features it cannot express, eroding linguistic semantics (question type decodability drops by $83\%$). By factorizing each modality into a shared core and a private fiber, D-JEPA allows private channels to carry modality-specific evidence without corrupting the common grounded core. The asymmetric independence penalty $\mathcal{L}_{\mathrm{ind}}$ further refines this separation when the underlying features are not already well-separated, as on CLEVR; on synthetic data, where the generative process already separates factors, the penalty is not required (\S\ref{sec:experiments}).

\paragraph{Choice of Diagnostic Benchmarks.}
We evaluate on controlled nonlinear manifolds and CLEVR precisely because these benchmarks expose the ground-truth latent factors ($S, U_1, U_2, Y_{\mathrm{syn}}$) and explicit relational query graphs that the PID audit and fiber-swap intervention require. The controlled setting is what makes the causal claims auditable: on CLEVR, the question taxonomy and scene graph allow us to compute exact per-category probe accuracy, and the synthetic benchmark gives known PID atoms against which the estimators can be checked. Together they provide the diagnostic resolution needed to distinguish shared, private, and synergistic contributions; natural web image-text pairs do not offer this resolution.

\paragraph{Limitations and Future Work.}
Our evaluation focuses on frozen ViT-S and BERT representations evaluated on CLEVR and synthetic manifolds at reduced scale. CLEVR provides rigorous ground truth for compositional question types, but scaling D-JEPA to web-scale pretraining datasets (e.g., DataComp, CC3M) and continuous robotic action spaces remains an important direction for future investigation. On CLEVR, D-JEPA narrows but does not close the gap to frozen-backbone baselines: joint D-JEPA accuracy is $39.61\%$, below raw text-only ($42.02\%$) and raw joint ($45.57\%$). The contribution is escaping the alignment bound, not recovering the full ceiling. On the synergy gap specifically, D-JEPA recovers $0.54$ of the raw $3.55$~pp on CLEVR ($15\%$); the near-complete recovery ($33.13$ of $35.00$~pp, $95\%$) is on the synthetic benchmark where the generative process isolates the factors. On CLEVR, text dominates because the question carries the relational operator, and vision contributes at the margin; the weaker recovery is the empirical form of the asymmetry the bound describes. Developing adaptive mechanisms that dynamically adjust capacity allocation between shared ($c$) and private ($u$) fibers based on mutual information density is a promising avenue for future work. Finally, the synthetic benchmark is evaluated over 2 seeds and CLEVR over 3; we report both and flag the asymmetry. A further direction is robustness under modality or sensor dropout: because the private fiber $\mathcal{U}_m$ is decorrelated from the shared core, we conjecture that D-JEPA would degrade more gracefully than alignment-only models when one sensory channel is lost; we leave a systematic evaluation to future work.

\paragraph{Conclusion.} An alignment-only representation is a projection, not a summary. While cross-modal alignment successfully recovers shared world structure in symmetric view regimes, it discards private and synergistic information under informational asymmetry. Decomposed JEPA resolves this limitation by preserving private representations via masked predictive grounding, closing the gap between alignment-only training and compositional multimodal understanding.

\bibliography{references}

@misc{radford2021clip,
      title={Learning Transferable Visual Models From Natural Language Supervision}, 
      author={Alec Radford and Jong Wook Kim and Chris Hallacy and Aditya Ramesh and Gabriel Goh and Sandhini Agarwal and Girish Sastry and Amanda Askell and Pamela Mishkin and Jack Clark and Gretchen Krueger and Ilya Sutskever},
      year={2021},
      eprint={2103.00020},
      archivePrefix={arXiv},
      primaryClass={cs.CV},
      url={https://arxiv.org/abs/2103.00020}, 
}

@misc{williams2010nonnegative,
      title={Nonnegative Decomposition of Multivariate Information}, 
      author={Paul L. Williams and Randall D. Beer},
      year={2010},
      eprint={1004.2515},
      archivePrefix={arXiv},
      primaryClass={cs.IT},
      url={https://arxiv.org/abs/1004.2515}, 
}

@article{bertschinger2014quantifying,
   title={Quantifying Unique Information},
   volume={16},
   ISSN={1099-4300},
   url={http://dx.doi.org/10.3390/e16042161},
   DOI={10.3390/e16042161},
   number={4},
   journal={Entropy},
   publisher={MDPI AG},
   author={Bertschinger, Nils and Rauh, Johannes and Olbrich, Eckehard and Jost, Jürgen and Ay, Nihat},
   year={2014},
   month=Apr, pages={2161–2183} }

@article{harder2013bivariate,
   title={Bivariate measure of redundant information},
   volume={87},
   ISSN={1550-2376},
   url={http://dx.doi.org/10.1103/PhysRevE.87.012130},
   DOI={10.1103/physreve.87.012130},
   number={1},
   journal={Physical Review E},
   publisher={American Physical Society (APS)},
   author={Harder, Malte and Salge, Christoph and Polani, Daniel},
   year={2013},
   month=Jan }

@misc{zhang2025flowpid,
      title={Partial Information Decomposition via Normalizing Flows in Latent Gaussian Distributions}, 
      author={Wenyuan Zhao and Adithya Balachandran and Chao Tian and Paul Pu Liang},
      year={2025},
      eprint={2510.04417},
      archivePrefix={arXiv},
      primaryClass={cs.LG},
      url={https://arxiv.org/abs/2510.04417}, 
}

@misc{bardes2022vicreg,
      title={VICReg: Variance-Invariance-Covariance Regularization for Self-Supervised Learning}, 
      author={Adrien Bardes and Jean Ponce and Yann LeCun},
      year={2022},
      eprint={2105.04906},
      archivePrefix={arXiv},
      primaryClass={cs.CV},
      url={https://arxiv.org/abs/2105.04906}, 
}

@misc{liang2022mind,
      title={Mind the Gap: Understanding the Modality Gap in Multi-modal Contrastive Representation Learning}, 
      author={Weixin Liang and Yuhui Zhang and Yongchan Kwon and Serena Yeung and James Zou},
      year={2022},
      eprint={2203.02053},
      archivePrefix={arXiv},
      primaryClass={cs.CL},
      url={https://arxiv.org/abs/2203.02053}, 
}

@inproceedings{andrew2013deep,
	title = {Deep Canonical Correlation Analysis},
	author = {Andrew, Galen and Arora, Raman and Bilmes, Jeff and Livescu,
	          Karen},
	booktitle = {Proceedings of the 30th International Conference on Machine
	             Learning (ICML)},
	series = {Proceedings of Machine Learning Research},
	volume = {28},
	pages = {1247--1255},
	publisher = {PMLR},
	year = {2013},
}

@misc{zbontar2021barlow,
      title={Barlow Twins: Self-Supervised Learning via Redundancy Reduction}, 
      author={Jure Zbontar and Li Jing and Ishan Misra and Yann LeCun and Stéphane Deny},
      year={2021},
      eprint={2103.03230},
      archivePrefix={arXiv},
      primaryClass={cs.CV},
      url={https://arxiv.org/abs/2103.03230}, 
}

@inproceedings{tsai2019learning,
	title = {Learning Factorized Multimodal Representations},
	author = {Tsai, Yao-Hung Hubert and Liang, Paul Pu and Zadeh, Amir and
	          Morency, Louis-Philippe and Salakhutdinov, Ruslan},
	booktitle = {International Conference on Learning Representations (ICLR)
	             },
	year = {2019},
}

@misc{thrush2022winoground,
	title = {Winoground: Probing Vision and Language Models for
	         Visio-Linguistic Compositionality},
	author = {Tristan Thrush and Ryan Jiang and Max Bartolo and Amanpreet
	          Singh and Adina Williams and Douwe Kiela and Candace Ross},
	year = {2022},
	eprint = {2204.03162},
	archivePrefix = {arXiv},
	primaryClass = {cs.CV},
	url = {https://arxiv.org/abs/2204.03162},
}

@misc{shi2019variational,
      title={Variational Mixture-of-Experts Autoencoders for Multi-Modal Deep Generative Models}, 
      author={Yuge Shi and N. Siddharth and Brooks Paige and Philip H. S. Torr},
      year={2019},
      eprint={1911.03393},
      archivePrefix={arXiv},
      primaryClass={stat.ML},
      url={https://arxiv.org/abs/1911.03393}, 
}

@misc{johnson2017clevr,
      title={CLEVR: A Diagnostic Dataset for Compositional Language and Elementary Visual Reasoning}, 
      author={Justin Johnson and Bharath Hariharan and Laurens van der Maaten and Li Fei-Fei and C. Lawrence Zitnick and Ross Girshick},
      year={2016},
      eprint={1612.06890},
      archivePrefix={arXiv},
      primaryClass={cs.CV},
      url={https://arxiv.org/abs/1612.06890}, 
}

@misc{jia2021align,
      title={Scaling Up Visual and Vision-Language Representation Learning With Noisy Text Supervision}, 
      author={Chao Jia and Yinfei Yang and Ye Xia and Yi-Ting Chen and Zarana Parekh and Hieu Pham and Quoc V. Le and Yunhsuan Sung and Zhen Li and Tom Duerig},
      year={2021},
      eprint={2102.05918},
      archivePrefix={arXiv},
      primaryClass={cs.CV},
      url={https://arxiv.org/abs/2102.05918}, 
}
\bibliographystyle{plainnat}

\newpage
\appendix
\section{Proofs}
\label{app:proofs}

\subsection{Proof of Theorem~\ref{thm:alignment_redundancy} and Corollary~\ref{cor:annihilation}}

\begin{proof}[Proof of Theorem~\ref{thm:alignment_redundancy}]
By Assumption~\ref{ass:markov}, the Markov chain $Z_1 - X_1 - S - X_2 - Z_2$ implies $Z_1 \perp Z_2 \mid S$, and therefore $I(Z_1; Z_2 \mid S) = 0$. Hence, the alignment term in~\eqref{eq:aib_obj} equals the three-way interaction information with $S$:
\begin{equation}
    I(Z_1; Z_2) = I(Z_1; Z_2) - I(Z_1; Z_2 \mid S) = I(S; Z_1; Z_2).
\end{equation}
Applying the Williams--Beer PID identities $I(S; Z_1, Z_2) = \Red(S; Z_1, Z_2) + \Unq(S; Z_1) + \Unq(S; Z_2) + \Syn(S; Z_1, Z_2)$ and $I(S; Z_i) = \Red(S; Z_1, Z_2) + \Unq(S; Z_i)$, the interaction information reduces to
\begin{equation}
\label{eq:align_is_red_minus_syn}
    I(Z_1; Z_2) = I(S; Z_1) + I(S; Z_2) - I(S; Z_1, Z_2) = \Red(S; Z_1, Z_2) - \Syn(S; Z_1, Z_2).
\end{equation}
Next, since $Z_i \perp S \mid X_i$ implies $I(Z_i; S \mid X_i) = 0$, expanding $I(Z_i; X_i, S)$ via the chain rule in both orders gives
\begin{equation}
\label{eq:cost_decomp}
    I(Z_i; X_i) = I(S; Z_i) + I(Z_i; X_i \mid S) = \Red(S; Z_1, Z_2) + \Unq(S; Z_i) + I(Z_i; X_i \mid S).
\end{equation}
Substituting~\eqref{eq:align_is_red_minus_syn} and~\eqref{eq:cost_decomp} into~\eqref{eq:aib_obj} and collecting terms yields~\eqref{eq:aib_pid_exact}. For $\beta \in (0, \tfrac{1}{2})$, only $\Red(S; Z_1, Z_2)$ carries a positive coefficient $(1-2\beta) > 0$, while $\Unq(S; Z_i)$, $\Syn(S; Z_1, Z_2)$, and $I(Z_i; X_i \mid S)$ all carry strictly negative coefficients. For $\beta \ge \tfrac{1}{2}$, every coefficient is $\le 0$, so maximizing $\mathcal{J}_{\mathrm{AIB}}$ drives all non-negative terms to zero.
\end{proof}

\begin{proof}[Proof of Corollary~\ref{cor:annihilation}]
If $I(Z_i; X_i \mid X_{\neq i}) = 0$, then $Z_i \perp X_i \mid X_{\neq i}$. Combined with $Z_i \perp X_{\neq i} \mid X_i$, by the G\'acs--K\"orner double-Markov property $Z_i$ is conditionally independent of $(X_1, X_2)$ given a common deterministic factor $K(X_1, X_2) = h_1(X_1) = h_2(X_2)$ a.s.\ (and if $Z_1 = Z_2$ a.s., $Z_1$ is directly such a common function). By Williams--Beer monotonicity and self-redundancy,
\[
    I(S; Z_1, Z_2) \;\le\; I(S; K(X_1, X_2)) \;=\; \Red(S; K, K) \;\le\; \Red(S; K, X_2) \;\le\; \Red(S; X_1, X_2).
\]
Subtracting $I(S; Z_1, Z_2)$ from the world-state PID identity~\eqref{eq:world_pid} yields the stated gap.
\end{proof}

\subsection{Proof of Theorem~\ref{thm:sufficiency}}

\begin{proof}
Since encoders act only on their own inputs, $(Z_1, Z_2) \perp S \mid (X_1, X_2)$, so $I(S; Z_1, Z_2 \mid X_1, X_2) = 0$. Expanding $I(S; X_1, X_2, Z_1, Z_2)$ via the chain rule in both orders gives
\begin{equation}
    I(S; X_1, X_2) + \underbrace{I(S; Z_1, Z_2 \mid X_1, X_2)}_{=\,0} = I(S; Z_1, Z_2) + I(S; X_1, X_2 \mid Z_1, Z_2),
\end{equation}
which proves~\eqref{eq:suff_gap}. For discrete $X_i$, $I(S; X_1, X_2 \mid Z_1, Z_2) \le H(X_1, X_2 \mid Z_1, Z_2) \le \sum_{i=1}^2 H(X_i \mid Z_i) = \sum_{i=1}^2 (H(X_i) - I(Z_i; X_i))$, yielding~\eqref{eq:discrete_gap_bound}; the bound for $T = h(S)$ follows by $T - S - (X_1, X_2) - (Z_1, Z_2)$ and the Data Processing Inequality.

For Part 2, $I(S; X_1, X_2 \mid Z_1, Z_2) = 0$ gives $I(S; Z_1, Z_2) = I(S; X_1, X_2)$, while $I(S; X_i \mid Z_i) = 0$ (together with $I(S; Z_i \mid X_i) = 0$) gives $I(S; Z_i) = I(S; X_i)$ and $p(s \mid z_i) = p(s \mid x_i)$ almost surely on the joint support. Any PID measure determined by the marginals $(S, X_i)$ (e.g., Williams--Beer $I_{\min}$ or BROJA) therefore satisfies $\Red(S; Z_1, Z_2) = \Red(S; X_1, X_2)$, fixing $\Unq(S; Z_i) = I(S; Z_i) - \Red$ and $\Syn(S; Z_1, Z_2) = I(S; Z_1, Z_2) - \sum_i I(S; Z_i) + \Red$ identically.
\end{proof}

\subsection{Proof of Proposition~\ref{prop:fusion_arch}}

\begin{proof}
For Part 1, since $F = \Phi_{\mathrm{attn}}(Z_1, Z_2)$ is a deterministic function of $(Z_1, Z_2)$ and local encoding forms the Markov chain $S - (X_1, X_2) - (Z_1, Z_2) - F$, the Data Processing Inequality immediately gives $I(S; F) \le I(S; Z_1, Z_2) \le I(S; X_1, X_2)$. If $(Z_1, Z_2)$ is an exact AIB optimum, Corollary~\ref{cor:annihilation} bounds $I(S; Z_1, Z_2) \le \Red(S; X_1, X_2)$, yielding $I(S; F) \le \Red(S; X_1, X_2)$.

For Part 2, because $W_Q Z_2$ and $W_K Z_1$ interact multiplicatively inside $\alpha(Z_2, Z_1)$ while $Z_2$ is preserved through the residual path and $W_V Z_1$ routes query-selected tokens, $\Phi_{\mathrm{attn}}$ does not enforce $Z_1 = Z_2$ and can represent conditional joint operators $\Phi_{Z_2}(Z_1)$ on uncollapsed $(Z_1, Z_2)$. Under Theorem~\ref{thm:sufficiency}, $(Z_1, Z_2)$ preserves all four PID atoms of $I(S; X_1, X_2)$, so an adequately parameterized $\Phi_{\mathrm{attn}}$ can extract $\Red, \Unq_1, \Unq_2,$ and $\Syn$ up to $I(S; Z_1, Z_2)$.

For Part 3, writing $\alpha(Z_2, Z_1) W_V Z_1 = \sum_{j \notin B} \alpha_{\cdot, j} W_V Z_{1,j} + \sum_{j \in B} \alpha_{\cdot, j} W_V Z_{1,j}$, whenever $\alpha_{\cdot, B} = 0$ a.s.\ the output $F$ depends on $Z_1$ strictly through $Z_{1, B^c}$ (and $Z_2$). Hence $F \perp Z_{1, B} \mid (Z_2, Z_{1, B^c})$, which implies $I(Z_{1,B}; F \mid Z_2, Z_{1, B^c}) = 0$.
\end{proof}

\subsection{Proof of Proposition~\ref{prop:constrained_lagrangian} and Equation~\eqref{eq:dropout_floor}}

\begin{proof}[Proof of Proposition~\ref{prop:constrained_lagrangian}]
Combining~\eqref{eq:encoder_pid} and~\eqref{eq:cost_decomp} gives
\[
    I(Z_i; X_i \mid X_{\neq i}) + I(Z_i; X_{\neq i}) = I(Z_i; X_i) = \Red(S; Z_1, Z_2) + \Unq(S; Z_i) + I(Z_i; X_i \mid S).
\]
Summing over $i \in \{1, 2\}$ and subtracting $(1+\beta)\sum_{i=1}^2 I(Z_i; X_i \mid S)$ cancels $+1$ on the nuisance term and yields~\eqref{eq:lagrangian_pid}.
\end{proof}

\begin{proof}[Proof of~\eqref{eq:dropout_floor}]
Since $Z_i = (C_i, U_i)$ satisfies $(C_i, U_i) \perp S \mid X_i$, we have $I(S; C_i, U_i \mid X_i) = 0$. Expanding $I(S; X_i, C_i, U_i)$ via the chain rule in both orders gives
\[
    I(S; C_i, U_i) = I(S; X_i) - I(S; X_i \mid C_i, U_i) = \Red(S; X_1, X_2) + \Unq(S; X_i) - \delta_i,
\]
using the PID marginal identity $I(S; X_i) = \Red(S; X_1, X_2) + \Unq(S; X_i)$ and $\delta_i := I(S; X_i \mid C_i, U_i)$.
\end{proof}

\section{Per-Condition Probe Accuracies}
\label{app:per_condition}

Table~\ref{tab:per_condition} reports the complete per-condition downstream probing accuracy across all representational conditions and targets on the synthetic benchmark (mean $\pm$ std over 2 seeds, \%).

\begin{table}[h]
\centering
\small
\caption{\textbf{Per-condition downstream probe accuracies on the synthetic benchmark} (test set, mean $\pm$ std, \%). Probing single-modality vision ($v$), text ($t$), joint stream ($joint$), and individual component blocks ($c, u, c_1, u_1, c_2, u_2$).}
\label{tab:per_condition}
\resizebox{\textwidth}{!}{%
\begin{tabular}{l ccc cc cc cc}
\toprule
\textbf{Arm / Target} & \textbf{v} & \textbf{t} & \textbf{joint} & \textbf{c} & \textbf{u} & \textbf{$c_1$} & \textbf{$u_1$} & \textbf{$c_2$} & \textbf{$u_2$} \\
\midrule
\multicolumn{10}{l}{\textbf{Target: Shared Factor ($S$)}} \\
Raw                  & $81.53 \pm 2.17$ & $78.07 \pm 0.28$ & $84.60 \pm 3.11$ & --- & --- & --- & --- & --- & --- \\
Alignment-only        & $79.47 \pm 2.83$ & $74.27 \pm 1.70$ & $81.07 \pm 2.07$ & --- & --- & --- & --- & --- & --- \\
Alignment-only + Rec  & $78.80 \pm 2.45$ & $74.93 \pm 0.57$ & $81.27 \pm 2.17$ & --- & --- & --- & --- & --- & --- \\
D-JEPA (Full)        & $79.80 \pm 2.17$ & $73.87 \pm 1.51$ & $79.40 \pm 2.36$ & $72.47 \pm 11.03$ & $66.87 \pm 5.75$ & $72.13 \pm 9.62$ & $69.67 \pm 0.28$ & $65.47 \pm 10.56$ & $64.07 \pm 7.45$ \\
D-JEPA (No-Indep)    & $79.40 \pm 2.17$ & $74.87 \pm 1.23$ & $83.00 \pm 1.79$ & $81.53 \pm 2.36$ & $83.47 \pm 0.94$ & $80.00 \pm 1.89$ & $80.07 \pm 1.23$ & $75.67 \pm 1.04$ & $76.33 \pm 0.47$ \\
D-JEPA (No-Rec)      & $78.07 \pm 1.98$ & $72.87 \pm 1.41$ & $78.27 \pm 2.83$ & $72.00 \pm 11.31$ & $67.13 \pm 7.64$ & $68.00 \pm 13.58$ & $67.20 \pm 3.21$ & $61.20 \pm 16.22$ & $61.87 \pm 7.35$ \\
\midrule
\multicolumn{10}{l}{\textbf{Target: Synergistic Interaction Factor ($Y_{\mathrm{syn}}$)}} \\
Raw                  & $47.53 \pm 0.85$ & $50.87 \pm 0.09$ & $85.87 \pm 0.19$ & --- & --- & --- & --- & --- & --- \\
Alignment-only        & $51.53 \pm 0.09$ & $48.20 \pm 1.23$ & $48.33 \pm 2.73$ & --- & --- & --- & --- & --- & --- \\
Alignment-only + Rec  & $51.40 \pm 1.41$ & $49.33 \pm 1.70$ & $49.73 \pm 0.94$ & --- & --- & --- & --- & --- & --- \\
D-JEPA (Full)        & $51.07 \pm 0.94$ & $48.40 \pm 1.89$ & $\mathbf{84.20 \pm 0.47}$ & $49.33 \pm 1.70$ & $84.27 \pm 0.57$ & $49.60 \pm 3.77$ & $51.33 \pm 2.64$ & $49.80 \pm 1.41$ & $48.80 \pm 0.94$ \\
\bottomrule
\end{tabular}%
}
\end{table}

\section{Hyperparameters}
\label{app:hyperparameters}

Table~\ref{tab:hyperparameters} summarizes training and evaluation configurations across all experiments.

\begin{table}[h]
\centering
\small
\resizebox{\textwidth}{!}{%
\begin{tabular}{l l l}
\toprule
\textbf{Category} & \textbf{Parameter} & \textbf{Value} \\
\midrule
Synthetic Benchmark & Samples ($N$) / views / latent factors & $3{,}000$ / $3$ views / $4$ factors ($d=64$) \\
                    & Steps / batch size & $300$ / $256$ \\
                    & Latent component dim $d_{\mathrm{comp}}$ / Total dim & $64$ / $256$ \\
CLEVR Benchmark     & Images / QA pairs per image / Total pairs & $3{,}000$ / $6$ / $12{,}000$ \\
                    & Pretraining steps / batch size / views & $3{,}000$ / $256$ / $3$ \\
                    & Encoders: Vision / Language & DINOv2 ViT-S/14 ($4 \times 4$) / BERT-base ($T=32$) \\
Architectures       & Encoder depth / heads / width ($d_{\mathrm{model}}$) & $2$ layers / $4$ heads / $256$ \\
Losses              & $\lambda$ (SIGReg) / $w_{\mathrm{inv}}, w_{\mathrm{cross}}, w_{\mathrm{rec}}$ & $0.05$ / $1.0$ each \\
                    & $w_{\mathrm{ind}}$ (decorrelation penalty) & $0.2$ \\
Audits              & Probe-PID / Flow-PID projection dim & $64$ \\
\bottomrule
\end{tabular}%
}
\caption{\textbf{Experimental hyperparameters across synthetic and real evaluations.}}
\label{tab:hyperparameters}
\end{table}

\section{Factor Decodability and Identification Matrices (Synthetic)}
\label{app:idmat}

Table~\ref{tab:idmat} reports the full held-out $R^2$ of regressing ground-truth target factors onto learned subspace blocks across arms on the synthetic benchmark.

\begin{table}[h]
\centering
\small
\caption{\textbf{Factor decodability per subspace on synthetic data} (held-out $R^2$, mean across seeds). Alignment-only eliminates cross-modal private decodability, whereas D-JEPA concentrates shared and private factors into their respective designed blocks.}
\label{tab:idmat}
\resizebox{\textwidth}{!}{%
\begin{tabular}{l ccc ccc ccc}
\toprule
 & \multicolumn{3}{c}{\textbf{Raw Backbones}} & \multicolumn{3}{c}{\textbf{Alignment-only}} & \multicolumn{3}{c}{\textbf{D-JEPA (Proposed)}} \\
\cmidrule(lr){2-4} \cmidrule(lr){5-7} \cmidrule(lr){8-10}
\textbf{Subspace} & \texttt{shared} & \texttt{priv\_v} & \texttt{priv\_t} & \texttt{shared} & \texttt{priv\_v} & \texttt{priv\_t} & \texttt{shared} & \texttt{priv\_v} & \texttt{priv\_t} \\
\midrule
Modality 1: $h_1 / z_1 / c_1$ & $0.961$ & $0.960$ & $-0.033$ & $0.897$ & $0.790$ & $-0.047$ & $0.821$ & $0.252$ & $-0.031$ \\
Modality 2: $h_2 / z_2 / c_2$ & $0.949$ & $-0.026$ & $0.948$ & $0.888$ & $-0.056$ & $0.787$ & $0.793$ & $-0.036$ & $0.171$ \\
Private Vis: $u_1$ & --- & --- & --- & --- & --- & --- & $0.257$ & $\mathbf{0.880}$ & $-0.035$ \\
Private Txt: $u_2$ & --- & --- & --- & --- & --- & --- & $0.245$ & $-0.039$ & $\mathbf{0.863}$ \\
\bottomrule
\end{tabular}%
}
\end{table}

\begin{figure}[h]
\centering
\includegraphics[width=0.85\textwidth]{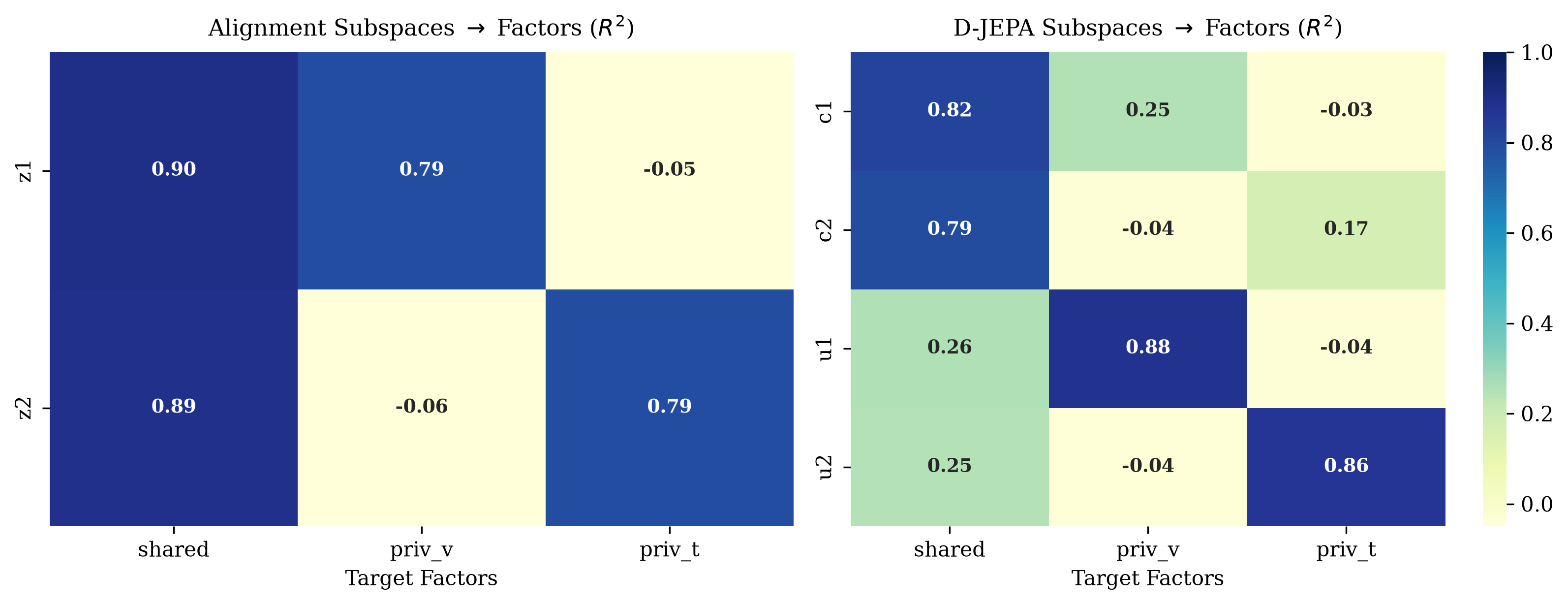}
\caption{\textbf{Feature dissection heatmap of learned subspaces ($R^2$) on synthetic data.} Decodability of ground-truth target factors from each subspace block across arms.}
\label{fig:factor_dissection}
\end{figure}

\section{Synthetic Information-Theoretic Audits}
\label{app:synthetic_pid}

Tables~\ref{tab:probe_pid_app} and~\ref{tab:flow_pid_app} report the Probe-PID and continuous Flow-PID audits on the synthetic benchmark.

\begin{table}[h]
\centering
\small
\caption{\textbf{Probe-PID decomposition on the synthetic benchmark} (bits; vision stream $S_1$, text stream $S_2$).}
\label{tab:probe_pid_app}
\resizebox{\textwidth}{!}{%
\begin{tabular}{l ccccc}
\toprule
\textbf{Model / Stream} & \textbf{Redundancy ($\Red$)} & \textbf{Vision-Unique ($\Unq_1$)} & \textbf{Text-Unique ($\Unq_2$)} & \textbf{Synergy ($\Syn$)} & \textbf{Total $I(Y; S_1, S_2)$} \\
\midrule
\multicolumn{6}{l}{\textbf{Target: Synergistic Interaction Factor ($Y_{\mathrm{syn}}$)}} \\
Raw ($h_1 \,\|\, h_2$)                   & $0.000$ & $0.000$ & $0.000$ & $\mathbf{0.528}$ & $\mathbf{0.528}$ bits \\
Alignment-only ($z_1 \,\|\, z_2$)         & $0.000$ & $0.000$ & $0.000$ & $0.000$ & $0.000$ bits \\
Alignment-only + Rec                 & $0.000$ & $0.000$ & $0.000$ & $0.000$ & $0.000$ bits \\
\textbf{D-JEPA (Full Stream)}            & $\mathbf{0.000}$ & $\mathbf{0.000}$ & $\mathbf{0.000}$ & $\mathbf{0.517}$ & $\mathbf{0.517}$ bits \\
D-JEPA (Shared Pair $[c_1\|c_2]$)         & $0.000$ & $0.000$ & $0.000$ & $0.000$ & $0.000$ bits \\
D-JEPA (Private Pair $[u_1\|u_2]$)       & $0.000$ & $0.000$ & $0.000$ & $0.496$ & $0.496$ bits \\
\midrule
\multicolumn{6}{l}{\textbf{Target: Shared Factor ($S$)}} \\
Raw ($h_1 \,\|\, h_2$)                   & $1.312$ & $0.078$ & $0.000$ & $0.101$ & $1.491$ bits \\
Alignment-only ($z_1 \,\|\, z_2$)         & $1.150$ & $0.134$ & $0.000$ & $0.054$ & $1.338$ bits \\
D-JEPA (Full Stream)                     & $1.130$ & $0.167$ & $0.000$ & $0.061$ & $1.359$ bits \\
\bottomrule
\end{tabular}%
}
\end{table}

\begin{table}[h]
\centering
\small
\caption{\textbf{Flow-PID continuous latent density decomposition on synthetic data} (bits; target: \texttt{shared}). Shuffled null baseline $I_{12}^{\mathrm{null}} \approx 0.12$ bits.}
\label{tab:flow_pid_app}
\resizebox{\textwidth}{!}{%
\begin{tabular}{l ccccc cc}
\toprule
\textbf{Model / Representation} & $\mathbf{I_1}$ & $\mathbf{I_2}$ & $\mathbf{I_{12}}$ & \textbf{Redundancy $\Red$} & $\mathbf{\Unq_1}$ & $\mathbf{Synergy \Syn}$ & $\mathbf{I_{12}^{\mathrm{null}}}$ \\
\midrule
Raw ($h_1 \,\|\, h_2$)                   & $2.170$ & $1.997$ & $\mathbf{2.644}$ & $1.997$ & $0.172$ & $0.474$ & $0.118$ \\
Alignment-only ($z_1 \,\|\, z_2$)        & $1.453$ & $1.512$ & $1.804$ & $1.453$ & $0.000$ & $0.292$ & $0.126$ \\
Alignment-only + Rec                & $1.473$ & $1.532$ & $1.816$ & $1.473$ & $0.000$ & $0.285$ & $0.117$ \\
\textbf{D-JEPA (Full Stream)}            & $1.506$ & $1.399$ & $\mathbf{1.838}$ & $1.399$ & $0.107$ & $\mathbf{0.332}$ & $0.125$ \\
D-JEPA (Shared Pair $[c_1\|c_2]$)         & $1.005$ & $0.905$ & $1.262$ & $0.905$ & $0.100$ & $0.257$ & $0.129$ \\
D-JEPA (Private Pair $[u_1\|u_2]$)       & $0.664$ & $0.626$ & $0.914$ & $0.626$ & $0.037$ & $0.250$ & $0.134$ \\
\bottomrule
\end{tabular}%
}
\end{table}

\section{Synthetic Causal Fiber-Swap Intervention}
\label{app:synthetic_swap}

Table~\ref{tab:fiber_swap_app} reports the randomized donor fiber-swap intervention on the synthetic benchmark.

\begin{table}[h]
\centering
\small
\caption{\textbf{Randomized donor fiber-swap intervention on the synthetic benchmark} (prediction flip rate, \%).}
\label{tab:fiber_swap_app}
\resizebox{\textwidth}{!}{%
\begin{tabular}{l cccc}
\toprule
\textbf{Target Task} & \textbf{Swap $c_1$ (Shared Vis)} & \textbf{Swap $u_1$ (Private Vis)} & \textbf{Swap $c_2$ (Shared Txt)} & \textbf{Swap $u_2$ (Private Txt)} \\
\midrule
Shared Target        & $33.8\%$ & $28.0\%$ & $30.2\%$ & $20.3\%$ \\
Private Vision       & $12.9\%$ & $\mathbf{78.0\%}$ & $10.7\%$ & $10.7\%$ \\
Private Text         & $13.0\%$ & $16.5\%$ & $14.5\%$ & $\mathbf{74.2\%}$ \\
Synergy Target       & $14.4\%$ & $\mathbf{48.8\%}$ & $14.6\%$ & $\mathbf{47.2\%}$ \\
\bottomrule
\end{tabular}%
}
\end{table}

\section{Cross-Modal Alignment Diagnostics}
\label{app:alignment}

Table~\ref{tab:alignment} reports cross-modal linear similarity (CKA), mutual nearest-neighbor overlap (mKNN), and Procrustes alignment on synthetic data, with a comparison against the CLEVR benchmark.

\begin{table}[h]
\centering
\small
\caption{\textbf{Cross-modal representation alignment diagnostics} (mean across seeds). Null = shuffled-pair baseline. On CLEVR, per-question text diversity means linear CKA and mKNN between modality streams are uninformative about functional alignment; downstream probe accuracy (\S4.2) and causal fiber-swap (\S4.6) are the operational tests.}
\label{tab:alignment}
\resizebox{\textwidth}{!}{%
\begin{tabular}{l cc cc cc}
\toprule
 & \multicolumn{2}{c}{\textbf{CKA}} & \multicolumn{2}{c}{\textbf{mKNN}} & \multicolumn{2}{c}{\textbf{Procrustes}} \\
\cmidrule(lr){2-3} \cmidrule(lr){4-5} \cmidrule(lr){6-7}
\textbf{Subspace Pair} & \textbf{Value} & \textbf{Null} & \textbf{Value} & \textbf{Null} & \textbf{Value} & \textbf{Null} \\
\midrule
\multicolumn{7}{l}{\textbf{Synthetic Ground-Truth Benchmark}} \\
Raw $h_1 \sim h_2$              & $0.232$ & $0.018$ & $0.045$ & $0.014$ & --- & --- \\
Alignment-only $z_1 \sim z_2$   & $0.914$ & $0.005$ & $0.394$ & $0.015$ & $0.938$ & $-0.002$ \\
Alignment-only + Rec $z_1 \sim z_2$ & $0.914$ & $0.006$ & $0.402$ & $0.015$ & $0.925$ & $-0.012$ \\
D-JEPA $c_1 \sim c_2$ (Shared)  & $0.867$ & $0.009$ & $0.281$ & $0.014$ & $0.892$ & $0.007$ \\
D-JEPA $u_1 \sim u_2$ (Private) & $0.091$ & $0.006$ & $0.024$ & $0.014$ & $0.123$ & $0.003$ \\
D-JEPA (No-Indep) $u_1 \sim u_2$& $0.432$ & $0.012$ & $0.105$ & $0.014$ & $0.471$ & $-0.006$ \\
\midrule
\multicolumn{7}{l}{\textbf{CLEVR Multimodal Benchmark}} \\
Raw $h_1 \sim h_2$              & $0.012$ & $0.006$ & $0.005$ & $0.004$ & --- & --- \\
Alignment-only $z_1 \sim z_2$   & $0.016$ & $0.004$ & $0.006$ & $0.005$ & $0.076$ & $0.003$ \\
Alignment-only + Rec $z_1 \sim z_2$ & $0.015$ & $0.004$ & $0.005$ & $0.004$ & $0.071$ & $0.003$ \\
D-JEPA $c_1 \sim c_2$ (Shared)  & $0.001$ & $0.001$ & $0.005$ & $0.005$ & $0.009$ & $0.001$ \\
D-JEPA $u_1 \sim u_2$ (Private) & $0.003$ & $0.003$ & $0.005$ & $0.005$ & $0.020$ & $-0.003$ \\
\bottomrule
\end{tabular}%
}
\end{table}

\end{document}